\documentclass[10pt,twocolumn,letterpaper]{article}
\usepackage[pdftex]{graphicx}
\usepackage[pagenumbers]{iccv}  %

\PassOptionsToPackage{table,dvipsnames}{xcolor}
\usepackage[table,dvipsnames]{xcolor}

\usepackage{colortbl}
\usepackage{xcolor}
\usepackage{graphicx}
\usepackage{algorithm}
\usepackage{algpseudocode}
\usepackage{symbols}

\usepackage{amsmath,amsfonts,bm}

\def\1{\bm{1}}

\def\sp{space}

\def\vc{{\bm{c}}}

\def\vx{{\bm{x}}}

\def\mA{{\bm{A}}}

\def\mI{{\bm{I}}}

\def\mM{{\bm{M}}}

\def\mV{{\bm{V}}}

\DeclareMathAlphabet{\mathsfit}{\encodingdefault}{\sfdefault}{m}{sl}
\SetMathAlphabet{\mathsfit}{bold}{\encodingdefault}{\sfdefault}{bx}{n}

\def\gC{{\mathcal{C}}}

\def\gN{{\mathcal{N}}}

\def\sR{{\mathbb{R}}}

\usepackage{mathtools}
\usepackage{listings}

\definecolor{codegreen}{rgb}{0,0.6,0}
\definecolor{codegray}{rgb}{0.5,0.5,0.5}
\definecolor{codepurple}{rgb}{0.58,0,0.82}
\definecolor{backcolour}{rgb}{0.95,0.95,0.92}

\lstdefinestyle{mystyle}{
    backgroundcolor=\color{backcolour},   
    commentstyle=\color{codegreen},
    keywordstyle=\color{magenta},
    numberstyle=\tiny\color{codegray},
    stringstyle=\color{codepurple},
    basicstyle=\ttfamily\footnotesize,
    breakatwhitespace=false,         
    breaklines=true,                 
    captionpos=b,                    
    keepspaces=true,                 
    numbers=left,                    
    numbersep=5pt,                  
    showspaces=false,                
    showstringspaces=false,
    showtabs=false,                  
    tabsize=2
}

\definecolor{darkred}{rgb}{0.7,0.1,0.1}
\definecolor{darkgreen}{rgb}{0.1,0.7,0.1}
\definecolor{cyan}{rgb}{0.7,0.0,0.7}
\definecolor{dblue}{rgb}{0.2,0.2,0.8}
\definecolor{maroon}{rgb}{0.76,.13,.28}
\definecolor{burntorange}{rgb}{0.81,.33,0}
\definecolor{tealblue}{rgb}{0.212,0.459, 0.533}

\definecolor{mypink}{rgb}{0.93359375, 0.62109375, 0.83984375}

\definecolor{pp}{rgb}{0.43921569, 0.18823529, 0.62745098}
\definecolor{rr}{rgb}{0.5254902 , 0.00784314, 0.12941176}
\definecolor{bb}{rgb}{0.09019608, 0.23529412, 0.37647059}
\definecolor{yy}{rgb}{0.49803922, 0.3372549 , 0.0}
\definecolor{gg}{rgb}{0.02352941, 0.3372549 , 0.17647059}

\definecolor{mygray}{rgb}{0.9, 0.9 , 0.9}
\usepackage{pifont}

\definecolor{iccvblue}{rgb}{0.21,0.49,0.74}
\usepackage[pagebackref,breaklinks,colorlinks,citecolor=iccvblue,linkcolor=lightcarminepink]{hyperref}
\definecolor{lightcarminepink}{rgb}{0.9, 0.4, 0.38}
\def\paperID{} %
\def\confName{}
\def\confYear{}

\title{
Tuning-Free Amodal Segmentation via \\the Occlusion-Free Bias of Inpainting Models
}

\author{Jae Joong Lee, Bedrich Benes, Raymond A. Yeh\\
Department of Computer Science, Purdue University\\
\{\small\tt lee2161, bbenes, rayyeh\}@purdue.edu}

\begin{document}
\maketitle
\setlength{\abovedisplayskip}{3pt}
\setlength{\belowdisplayskip}{3pt}
 \begin{abstract}
Amodal segmentation aims to predict segmentation masks for both the visible and occluded regions of an object. Most existing works formulate this as a supervised learning problem, requiring manually annotated amodal masks or synthetic training data. Consequently, their performance depends on the quality of the datasets, which often lack diversity and scale. This work introduces a tuning-free approach that repurposes pretrained diffusion-based inpainting models for amodal segmentation. Our approach is motivated by the ``occlusion-free bias'' of inpainting models, i.e., the inpainted objects tend to be complete objects without occlusions. Specifically, we reconstruct the occluded regions of an object via inpainting and then apply segmentation, all without additional training or fine-tuning. Experiments on five datasets demonstrate the generalizability and robustness of our approach. On average, our approach achieves 5.3\% more accurate masks over the state-of-the-art. 
\end{abstract}

 \section{Introduction}

Amodal segmentation refers to predicting segmentation masks even under occlusions~\cite{li2016amodal}. This challenging task involves reasoning about the unseen portion of an object under complex occlusion and illumination scenarios. It is an important problem with potential applications in autonomous driving and robot planning, which require reasoning beyond what is directly observed to predict possible future events in the environment~\cite{yang2019embodied, kins_original, dang2019graph}.

\begin{figure}[ht]
    \centering
    \small
    \setlength{\tabcolsep}{0.5pt} %
    \renewcommand{\arraystretch}{0.5} %

    \begin{tabular}{cccc} %
        Input & \multicolumn{3}{c}{Inpainted samples}\\
        
        \includegraphics[width=0.245\linewidth]{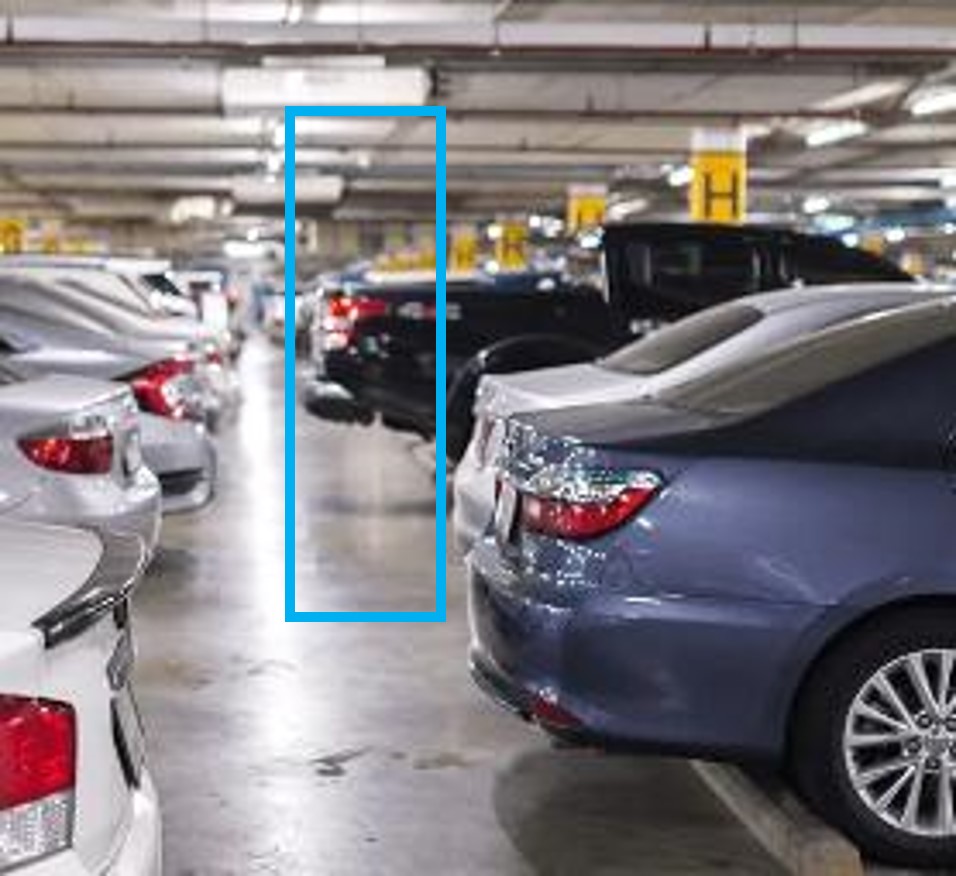} &
        \includegraphics[width=0.245\linewidth]{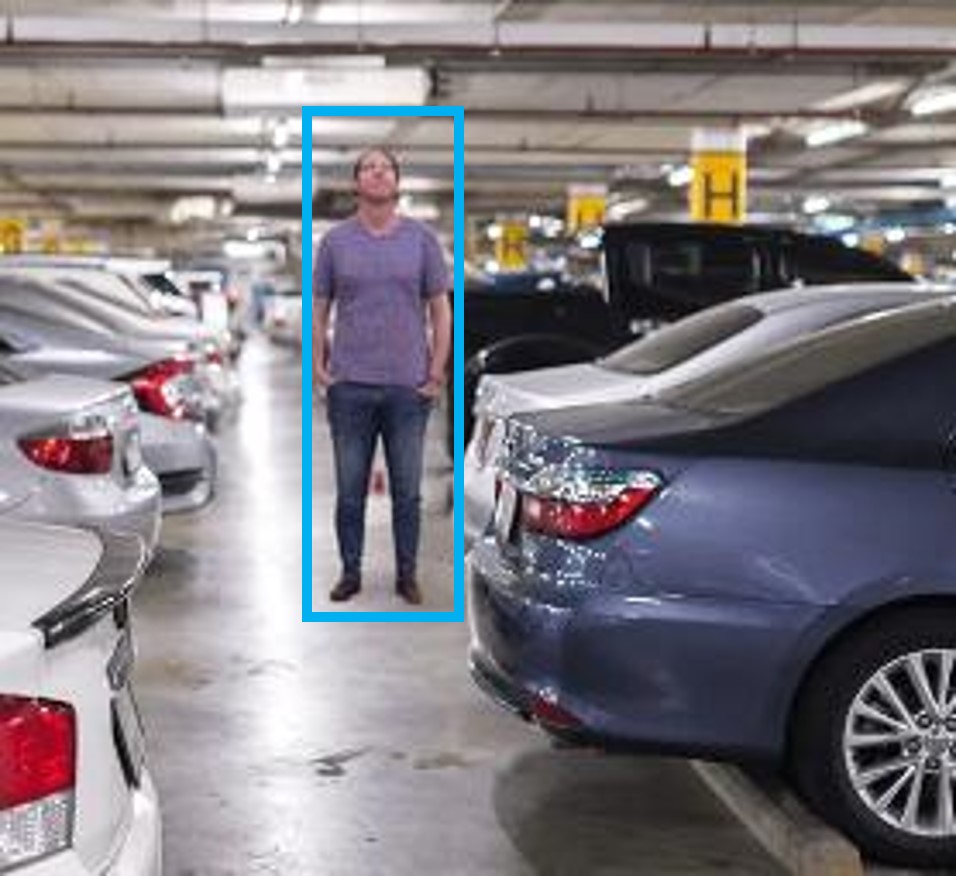} &
        \includegraphics[width=0.245\linewidth]{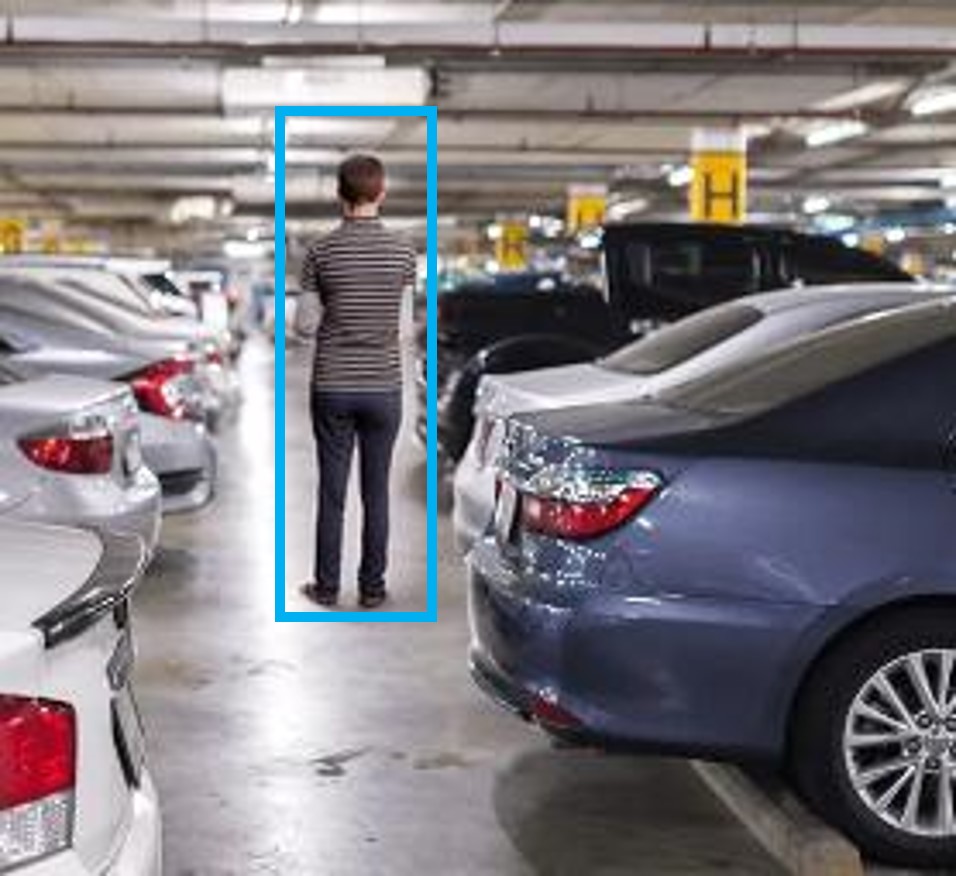} &
        \includegraphics[width=0.245\linewidth]{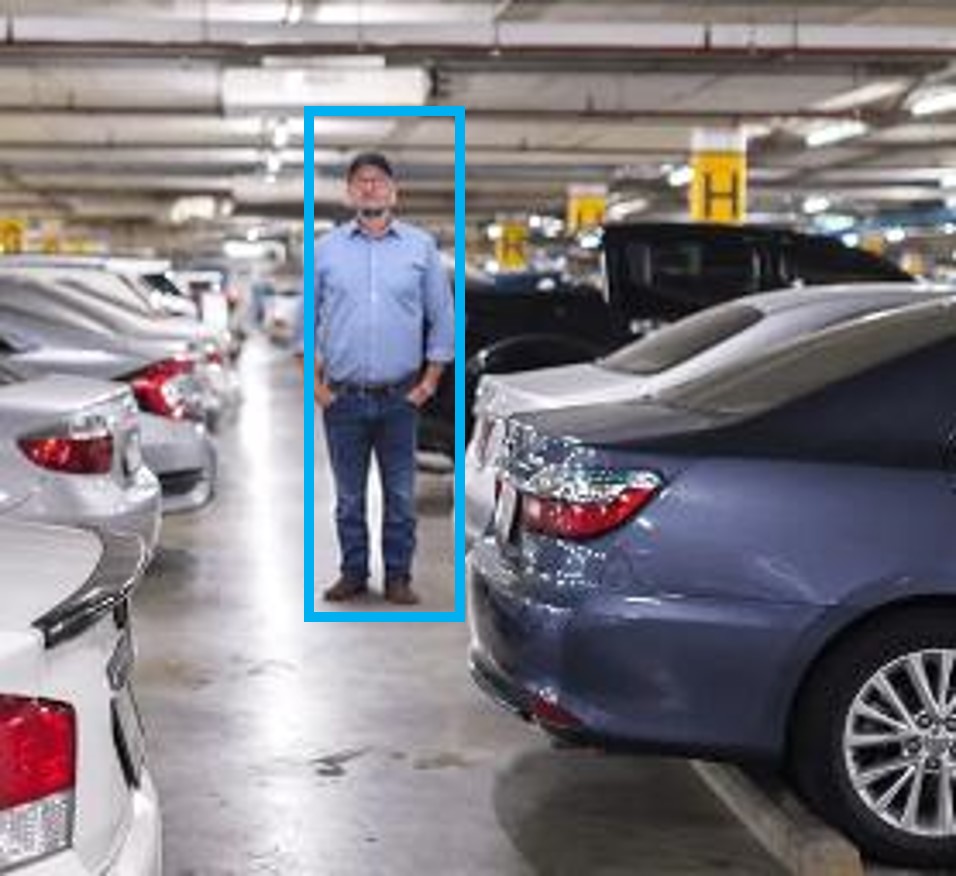} \\

        \includegraphics[width=0.245\linewidth]{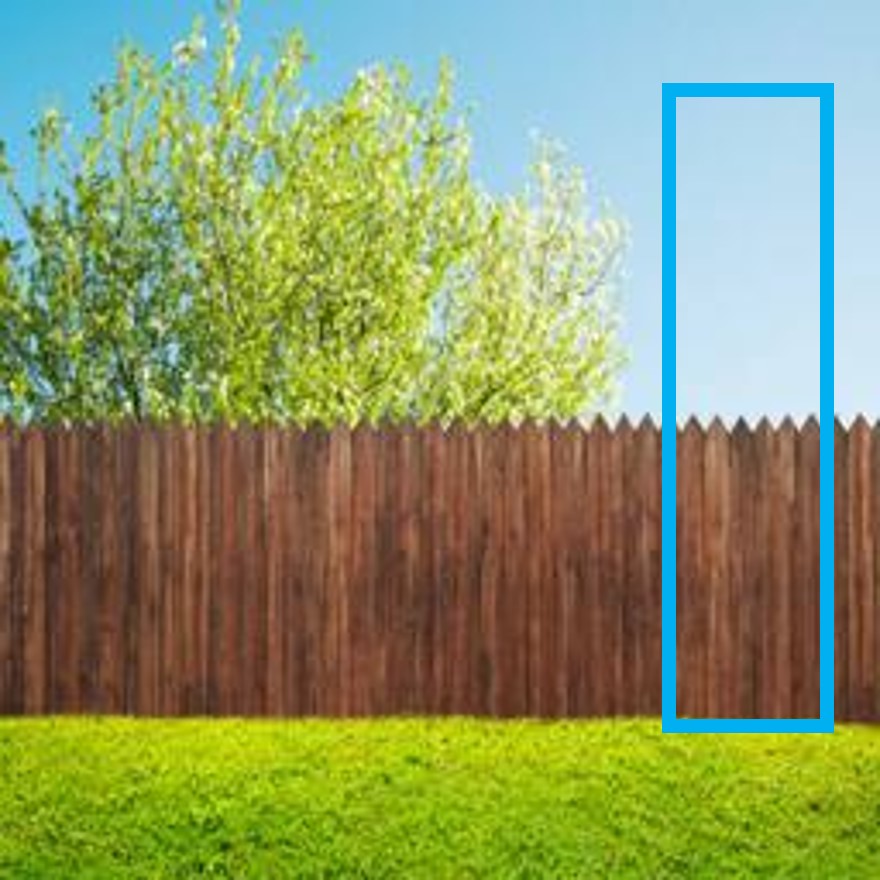} &
        \includegraphics[width=0.245\linewidth]{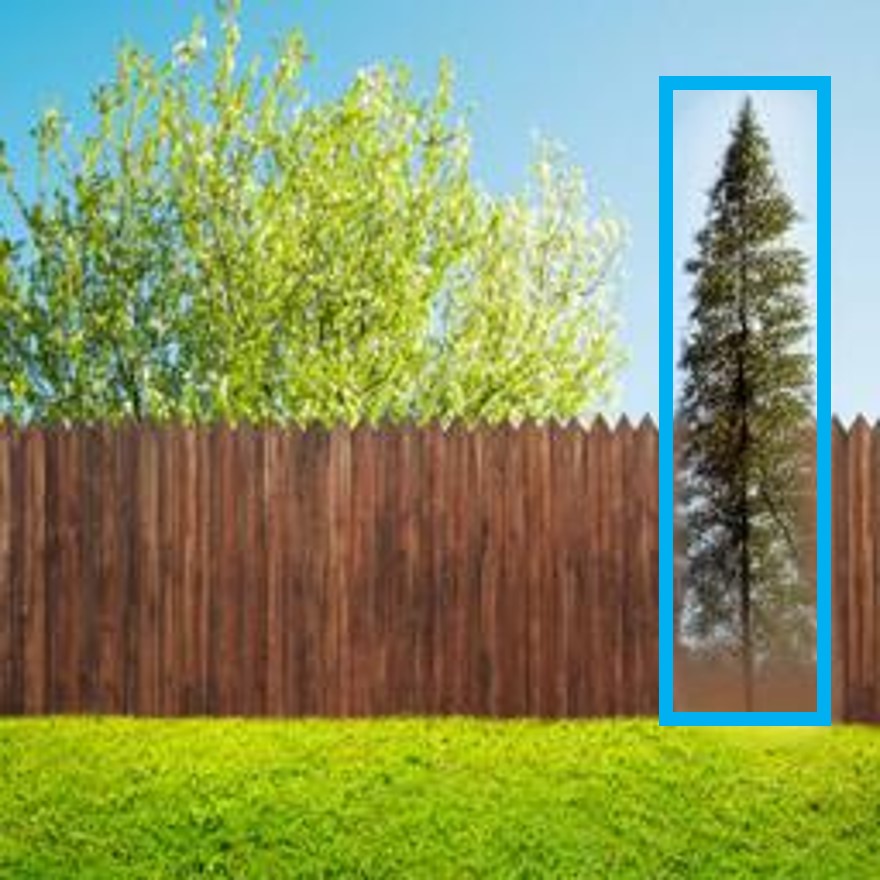} &
        \includegraphics[width=0.245\linewidth]{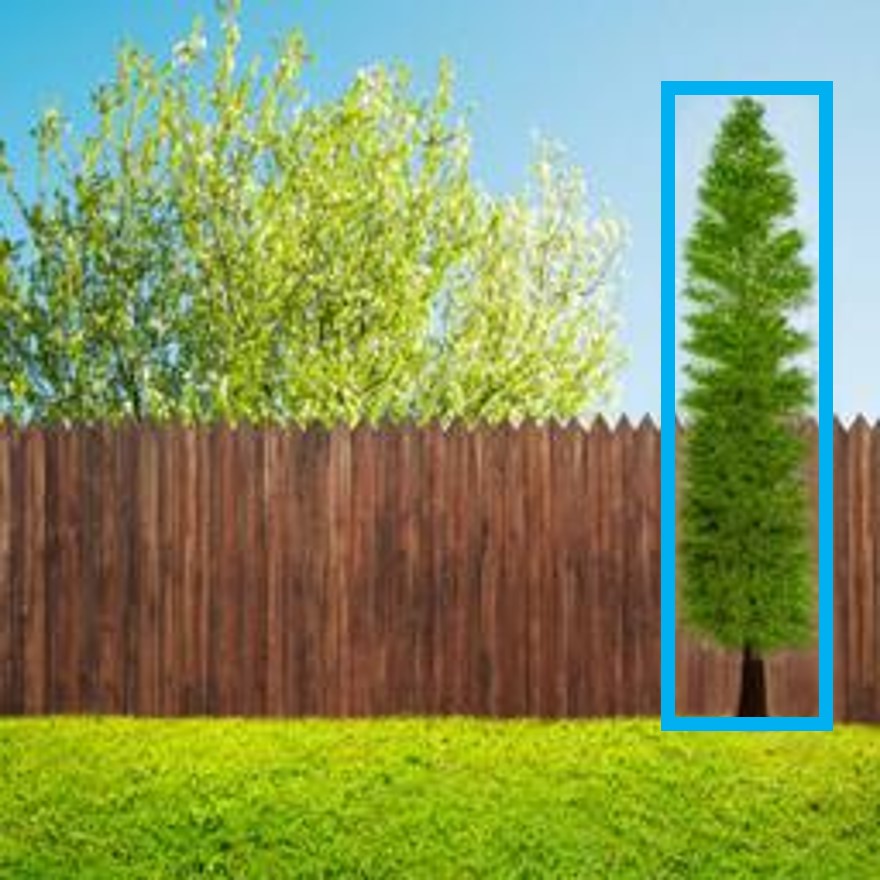} &
        \includegraphics[width=0.245\linewidth]{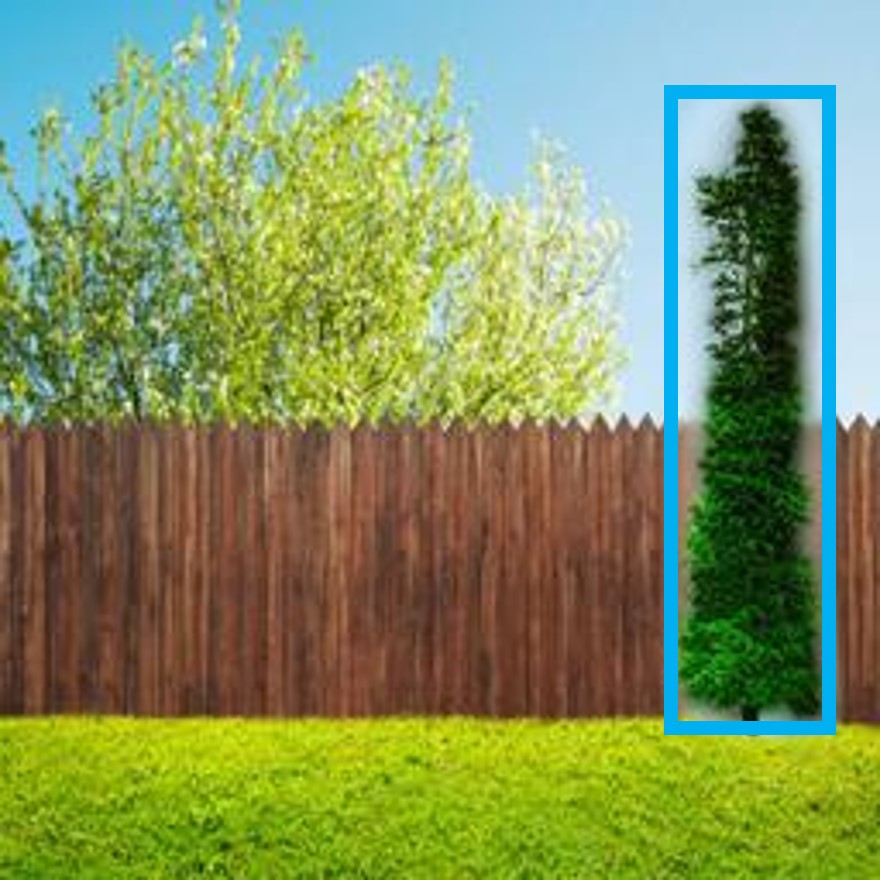} \\

    \end{tabular}
        \vspace{-0.25cm}
    \caption{{\bf Occlusion-free bias for a diffusion inpainting model.} We observe that an inpainted object is always placed without occlusions inside the inpainting area (blue box), \eg, a tree could have been inpainted behind the fence. %
    }
    \label{fig:occlusion_free_bias}
    \vspace{-0.5cm}
\end{figure}

Following the success of deep segmentation methods~\cite{sam, dino, long2015fully, ronneberger2015u}, amodal segmentation is often formulated as a supervised learning task, \ie, a dataset of (image, amodal mask) pairs is collected to train a model. However, preparing a large dataset for amodal segmentation is challenging. Annotating amodal masks requires reasoning over occluded reason, which may be difficult and inconsistent among human annotators. Furthermore, scaling the diversity of the dataset is challenging as it requires numerous combinations of occluders and objects.

Inevitably, several amodal segmentation methods turns to synthetically generating occlusions~\cite{xiao2021amodal, follmann2019learning, ozguroglu2024pix2gestalt, ao2024amodal} and %
3D game engine rendering~\cite{SAILVOS} to obtain the annotations. However, the performance is still limited by (a) the distribution gap between the synthetic and real data and (b) the size of the dataset. For example, the currently available SOTA~\cite{ozguroglu2024pix2gestalt} uses only 800k data pairs for training/fine-tuning, which is relatively small compared to the recent internet-scale datasets for other tasks~\cite{Rombach_2022_CVPR, peebles2023scalable, podell2023sdxl}. Another work~\cite{progressive} also proposes to use a pre-trained diffusion model, but it requires iterative occlusion removal and is constrained to a fixed 83 object categories, which has limited generalizability.

To address these challenges, we present a \textit{tuning-free approach} that utilizes existing foundation models trained on internet-scale datasets. Our method \textbf{does not require} any amodal data. Hence, the method is naturally zero-shot and without restriction to pre-defined object classes. Our approach is motivated by the observation that diffusion inpainting models have an ``occlusion-free bias'', \ie, the inpainting model prefers to generate a whole object rather than the occluder given a reasonable mask as shown in~\figref{fig:occlusion_free_bias}. %
We propose to perform inpainting over an enlarged modal mask, where the diffusion model fills the occluded regions. With the inpainted occluded regions, we extract the modal segmentation as the amodal prediction.  

\begin{figure*}[t]
    \centering
    \includegraphics[width=0.97\linewidth]
    {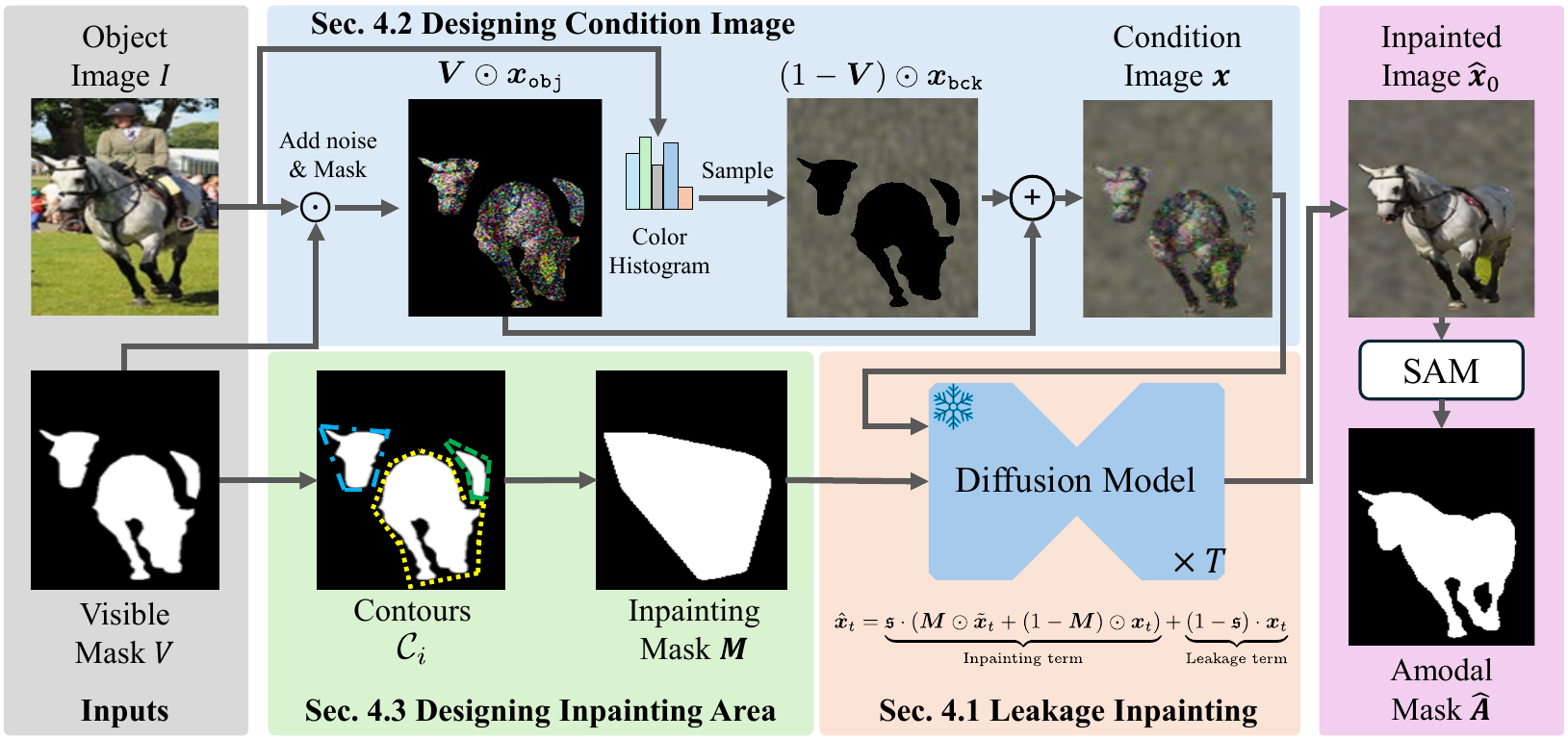}
    \vspace{-0.2cm}
    \caption{Our approach takes two inputs an RGB image  $\mI$  and a visible mask  $\mV$. From $\mI$, we generate a conditioned RGB image with a color distribution-aware background $\vx_{\tt bck}$ and a partial Gaussian noise-added object $\vx_{\tt obj}$. From $\mV$, we create a customized inpainting area $\mM$ so that we utilize any diffusion-based inpainting models to create an inpainted image $\hat\vx_0 $ to extract amodal mask $\hat\mA$. %
    }
    \vspace{-0.35cm}
    \label{fig:overview}
\end{figure*}

We demonstrate the effectiveness of our approach on five diverse amodal segmentation datasets, namely, COCO-A~\cite{cocoa}, BSDS-A~\cite{bsdsa}, KINS~\cite{kins}, FishBowl~\cite{fishbowl}, and SAILVOS~\cite{SAILVOS}. Notably, our tuning-free approach outperforms the current supervised SOTA~\cite{ozguroglu2024pix2gestalt} by an average of 5.3\% in the mIoU. %

{\bf\noindent Our contributions are as follows:}
\begin{itemize}
    \item We propose a tuning-free method for amodal segmentation (zero-shot) by exploiting the occlusion-free bias of diffusion inpainting models.
    \item The method involves several novel components, including a context-aware approach to background composition using RGB distribution, a noising process image for conditioning, and a modal mask construction procedure.
    \item We demonstrate the generalizability of the proposed method by conducting extensive experiments over four diffusion inpainting models on five diverse datasets.
\end{itemize}

 \section{Related Work}
{\bf\noindent Amodal perception and segmentation.}
Humans can {often} detect and identify an object even if it is {(partially)} occluded~\cite{lehar1999gestalt}. Seminal work by~\citet{li2016amodal} begins the line of work of using deep learning for amodal tasks. Many architectures and models have been proposed, \eg, CNN~\cite{li2016amodal,zhang2019learning, yang2019embodied,xiao2021amodal}, Generative Adversarial Networks~\cite{ehsani2018segan}, Transformer~\cite{tran2022aisformer, gao2023coarse}, and Diffusion-based models~\cite{ozguroglu2024pix2gestalt, zhan2024amodal}. Existing works are evaluated on datasets such as COCO-A~\cite{cocoa}, BSDS-A~\cite{bsdsa}, KINS~\cite{kins}, and MP3D-Amodal~\cite{zhan2024amodal}, which consist of common objects from the real world. Other synthetic benchmarks are also popular, \eg, SAILVOS~\cite{SAILVOS} or FishBowl~\cite{fishbowl}. These synthetic datasets provide more diverse object categories and precise amodal mask annotations without human errors. %

The current SOTA in amodal segmnetation that involves training are pix2gestalt~\cite{ozguroglu2024pix2gestalt} and AmodalWild~\cite{zhan2024amodal}. We refer to pix2gestalt as the SOTA as its code is ultimately released, whereas AmodalWild did not release the training code. pix2gestalt~\cite{ozguroglu2024pix2gestalt}, which trains a deep-net to predict the occluded pixels following an analysis by synthesis framework~\cite{yuille2006vision}. Specifically, they created a synthetic amodal dataset by occluding objects with randomly sampled overlays using another object. Our work does not require any amodal datasets, \ie, it is tuning-free.

Next, the closest related to our work is the tuning-free method proposed by~\citet{progressive}, where they propose to iteratively use an inpainting modal for amodal completion, \ie, they are interested in high image quality. Nonetheless, as a modal mask can be extracted from the completed image, we consider it to be an amodal segmentation method. The key limitation of~\citet{progressive}, upon reviewing their code~\cite{k8xu_amodal_classes}, is that the approach leverages class information and is limited to only 83 classes %
his greatly limits the usability of their approach. In contrast, our method does not have class restrictions and does not require multiple calls to the inpainting model.

{\bf\noindent Image inpainting} is the task of filling in missing regions of a given image, where the missing region is indicated using a mask. Early works in inpainting leverage low-level properties of natural images, \eg, smoothness~\cite{tschumperle2005vector, darabi2012image} or low-rank~\cite{jin2015annihilating, guo2017patch}, to tackle this task. In cases where the image contains a large missing region, then generative or learning-based methods are proposed~\cite{yeh2017semantic, yu2019free, zeng2020high, li2020recurrent, guo2021image, li2022mat, liu2022partial, yildirim2023diverse}.
More recently, diffusion models have emerged as the state-of-the-art in image generation, naturally, image inpainting methods based on diffusion have also been proposed~\cite{lugmayr2022repaint, suvorov2022resolution,Rombach_2022_CVPR,saharia2022palette,liu2024prefpaint,corneanu2024latentpaint}. 
This work leverages pre-trained diffusion inpainting models for the tasks of zero-shot amodal segmentation, \ie, we do not require high image quality, only accurate object contours.

{\bf\noindent Tuning-free methods for diffusion models.}
A diffusion model requires training on a large number of images to generate a realistic output. 
Although pre-trained diffusion models exist, fine-tuning is still needed for new tasks. Recent tuning-free methods leverage pre-trained models for performance gains and other tasks without extra training, improving Text-to-Image synthesis~\cite{zeng2024jedi, ding2024freecustom} and video generation~\cite{qiu2024freenoise, he2023scalecrafter}.
Our approach aligns with these works as we do not require additional fine-tuning. Differently, this paper focuses on using pre-trained inpainting models for the {\it task of amodal segmentation}, and the method is tuning-free.

 \section{Preliminaries}
We review diffusion models~\cite{ddpm} and inpainting with diffusion~\cite{lugmayr2022repaint}.
Diffusion models add noise to the data (forward process) and learn to undo the added noise (reverse process) during training. At generation, diffusion models start from a purely sampled noise and perform the reverse process. %

{\bf{\noindent Forward diffusion process}} gradually adds Gaussian noise to a clean image, $\vx_0$, over $T$ timesteps where $\vx_t$ is the noisy version of the image at timestep $t$ with  $\alpha_t$ controlling the amount of noise added at each step. The noisy image $x_t$ can be computed from $x_0$ as follows
\begin{align}
\label{eqn:diffusion_forward}
\vx_t &= \sqrt{\bar{\alpha}_t} \vx_0 + \sqrt{1 - \bar{\alpha}_t} \epsilon
\end{align}
where $ \bar{\alpha}_t = \prod_{s=1}^t \alpha_s $ is the cumulative product of the noise scaling factors and $ \epsilon \sim \mathcal{N}(0, I) $ is a Gaussian noise.

{\bf{\noindent Reverse diffusion process}} undoes the forward diffusion by denoising an image iteratively, starting from a pure noise image, $\vx_T$ to the clean image, $\vx_0$. This is formulated as a sequence of conditional probabilities 
\begin{align}
\label{eqn:diffusion_reverse}
p_\theta(\vx_{t-1} \mid \vx_t) &= \mathcal{N}(\vx_{t-1}; \mu_\theta(\vx_t, t), \sigma_t\mI)
\end{align}
following the Gaussian distribution with mean $ \mu_\theta(\vx_t, t) $, and diagonal covariance matrix predicted from a deep-net with parameters $\theta$. Intuitively, $\mu_\theta$ can be thought of as acting as an image denoiser that gradually removes noise according to a schedule. Another common choice, introduced by DDPM~\cite{ddpm}, is to use a deep-net to model the residual noise $\epsilon_\theta$. This is equivalent to choosing a denoiser 
\bea
\mu(\vx,t) \triangleq \vx - \sigma_t \cdot \epsilon_\theta(\vx, t)
\eea
{\bf{\noindent Inpainting}} predicts masked-out regions of a given input image. Diffusion-based method~\cite{lugmayr2022repaint} leverages the generative prior of a pre-trained DDPM~\cite{ddpm} to do so. This is achieved by iteratively removing noise from the linear combination of the noisy unmasked regions with the generated mask regions. More formally, given an input image $\vx$ and a mask $\mM \in \{0,1\}^{H\times W}$ the generation process to produce an inpainted image $\hat\vx_0$ is as follows:
\bea\label{eq:in_sample}
\tilde\vx_{t} & \sim & \gN\left(\mu_\theta(\hat\vx_{t-1}, t) , \sigma_{t-1})\right)\\
\label{eq:in_mask}
\hat\vx_{t} & = & \mM \odot \tilde\vx_{t} + (1-\mM) \odot \vx_t
\eea
where $\vx_t$ is the noise added input image following~\equref{eqn:diffusion_forward}, $\tilde\vx_T$ is assumed to be pure noise, and $\odot$ denotes element-wise multiplication. Recent foundation diffusion models~\cite{Rombach_2022_CVPR, podell2023sdxl,flux} are text-conditioned. The denoiser takes in an additional text prompt $\vc$ to guide the generation, \ie, 
\bea
\tilde\vx_t(\vc) \sim \mu_\theta\left(\hat\vx_{t-1}, \vc, t\right).
\eea

 \section{Training-free Amodal Segmentation}
{\noindent\bf Problem formulation.}
We consider the amodal segmentation setup as in~\citet{ozguroglu2024pix2gestalt}. Given an object's image $\mI$ and corresponding visible (modal) mask $\mV$, the task is to predict the object's amodal mask $\hat\mA$ that covers the whole object, including occluded regions. 

{\noindent\bf Overview.} We propose a tuning-free method for amodal segmentation by re-purposing diffusion inpainting models. Our approach leverages the ``occlusion-free'' bias of diffusion inpainting models, as shown in~\figref{fig:occlusion_free_bias}, where an inpainted object is almost always generated without occlusion. Hence, we inpaint an occluded object to remove the occlusion and use a segmentation method, \eg, SAM~\cite{sam}, on the unoccluded object to extract the amodal mask $\hat\mA$. While the proposed method seems straightforward, the devil is in the details. 

To achieve high-quality amodal masks, we needed to carefully design the generation procedure of the inpainting model (\secref{sec:leakage_cond}), the conditioning image $\vx$ (\secref{sec:cond_img}), and the inpainting area $\mM$ (\secref{sec:inpaint_area}). A visual illustration of the approach is provided in~\figref{fig:overview}.

\begin{figure*}[t]
    \centering
    \small
    \includegraphics[width=0.99\linewidth]
    {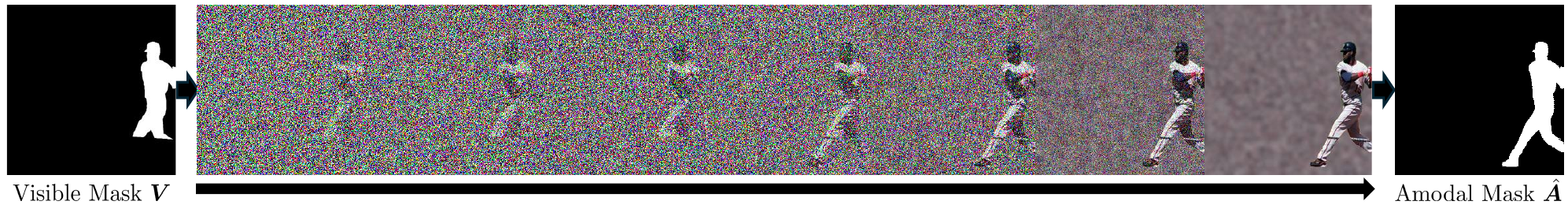}
    \vspace{-0.15cm}
    \caption{We show a visual process of the diffusion model. As we are doing soft-inpainting, observe that our approach can predict an amodal mask much larger than the visible mask, \ie, extrapolate.
    }
    \vspace{-0.35cm}
    \label{fig:extrapolation_process}
\end{figure*}

\subsection{Inpainting via leakage conditioning }
\label{sec:leakage_cond}
Recent diffusion inpainting models~\cite{Rombach_2022_CVPR, podell2023sdxl, flux} are often text-conditioned, \ie, the model performs a conditional generation on the masked area with a text-prompt.
As the task of amodal segmentation does not involve any text prompt, we need another way to condition the model. Specifically, besides the standard diffusion sampling for inpainting we further ``leak'' the original unmasked conditioning image $\vx$ to the model. Instead of~\equref{eq:in_mask}, we perform the following:
\begin{align}
\label{eqn:noise_control}
\hat\vx_{t} = \underbrace{\mathfrak{s} \cdot \left(\mM \odot \tilde\vx_{t} + (1-\mM\right) \odot \vx_t)}_{\text{Inpainting term}} + \underbrace{(1 - \mathfrak{s}) \cdot \vx_t}_{\text{Leakage term}},
\end{align}
where $\mathfrak{s} \in \sR^+$ controls the strength of the leakage. 
The purpose of the leakage term is that we want the model to inpaint occluded parts relevant to the current scene context, both masked and non-masked regions. Empirically, we set $\mathfrak{s}=0.3$ to balance the level of image context preservation, which is equivalent to using an image that is the combination of 30\% of a newly generated image and 70\% of the original image to maintain the overall original context. Increasing noise loses visual contexts, leading to random objects with poor quality of amodal segmentation.

Please note that the update equation in~\equref{eqn:noise_control} \textit{no longer strictly} performs image inpainting as the non-masked region is {\it not} guaranteed to be the same as the conditioning image $\vx$. Instead, we perform a ``soft''-inpainting where the model generates an image that roughly resembles the condition image $\vx$ for the unmasked regions and focuses on generating within the inpainting area. As the inpainting area is not strict, this also has the benefit that {\bf pixels outside of the inpainting area $\mM$ can be changed},~\ie, the predicted amodal mask can be larger than the inpainting area $\mM$, which helps with cases where extrapolation of the visible mask is needed. 
To handle this, we leverage the leakage from~\cref{eqn:noise_control}, which acts like a ``soft scaffold'', allowing the model to perform ``soft-inpainting'', changing pixels outside the mask. We show this process visually in~\cref{fig:extrapolation_process} and more extrapolation examples in~\cref{fig:extrapolation}.

\subsection{Designing the condition image}
\label{sec:cond_img}
The inpainting procedure needs an input condition image, denoted as $\vx$, to guide the generation process. Our objective is to create a complete object without any occlusions. To achieve this, we want the model to focus on the visible parts of the object rather than the background. Therefore, we have a separate procedure for preparing the object and background pixels, where 
\bea
\vx = \mV \odot \vx_{\tt obj} + (1-\mV) \odot \vx_{\tt bck}.
\eea

{\bf\noindent Object pixels.} Given the image $\mI$ containing the object and its corresponding visible mask $\mV$, we extract the object pixels by overlaying the visible mask using an element-wise multiplication. Next, as a diffusion model expected a noisy image, we add noise to the object pixels similar to~\cite{corneanu2024latentpaint, meng2022sdedit}, \ie,
\bea
\vx_{\tt obj} =  
\left(\mathfrak{s} \cdot \epsilon + (1-\mathfrak{s}) \cdot \mI\right),
\eea
where $\epsilon \sim \gN(0,\mI)$ and $\mathfrak{s}=0.3$.

{\bf\noindent Background pixels.}
In standard inpainting, the pixel values in the masked-out region (background) do not matter. Hence, it is common to choose either black or white color. However, the background now plays a role due to our leakage conditioning in~\equref{eqn:noise_control}. The default choice of black or white introduces a sharp contrast around the object's contours, and diffusion models do not react well to this pixel intensity discontinuity. 

Inspired by previous works to blend images seamlessly, such as leveraging the denoising process~\cite{lugmayr2022repaint} and incorporating latent information from text-guided diffusion models~\cite{avrahami2022blended}. We construct a smooth background that matches the object's color distribution. First, we build a color histogram from the object's visible pixels in $\mI$, then sample background pixels $\vx_{\tt{bck}}$ based on histogram frequencies, and finally apply a Gaussian blur.

\subsection{Designing the soft inpainting area}
\label{sec:inpaint_area}
Besides the condition image $\vx$, the inpainting procedure also requires an inpainting area $\mM$, which specifies where to focus on the generation. 

From the visible mask $\mV$, we extract contours from a set of points corresponding to visible regions, where \( \gC_{i} \) is the $i^{\text{th}}$ contour. Next, we combine the contours into one region by taking their union and finding the smallest convex polygon $\texttt{CnvxHull}(\bigcup_{i=1} \gC_i)$ that can enclose all contours.
Finally, we get the inpainting region, $\mM$, by setting values inside the convex polygon to one, where a visible pixel at $\left( x, y \right)$:
\bea
\mM =
\begin{cases}
1, & \text{if } (x, y) \in \texttt{CnvxHull}(\bigcup_{i=1} \gC_i), \\
0, & \text{otherwise}.
\end{cases}
\eea
Next, diffusion inpainting models are trained to take an inpainting region as conditioning input, \ie, $\tilde\vx_t(\mM)$. Hence, this allows us to use classifier-free guidance~\cite{ho2022classifier} with the mask during the generation. Let $w$ denote the intensity of the strictness which is how the model needs to follow the conditioning of the inpainting region $\mM$ a classifier-free-guided sample computes
\bea
\tilde\vx^{\tt{CFG}}_{t} = (1 + w) \cdot \tilde{\vx}_{t} (\mM) - w \cdot \tilde\vx_{t} (\varnothing) , \label{eqn:cfg}
\eea
where $\varnothing$ denotes the empty representation of $\mM$.
Instead of using $\tilde\vx$ directly in~\equref{eq:in_mask}, $\tilde\vx^{\tt{CFG}}_{t}$ is used. 
Intuitively, smaller~$w$ gives more freedom to generate new pixel information independent of the inpainting area shape. Empirically, we set $w=0.75$ for Stable Diffusion version 1.5~\cite{Rombach_2022_CVPR},  Stable Diffusion XL~\cite{podell2023sdxl} and $w$ as 1.5 with Flux~\cite{flux}.

\begin{table}[t]
    \small
    \setlength{\tabcolsep}{2pt}
    \centering
    \resizebox{\linewidth}{!}{%
    \begin{tabular}{ccccccc|c}
    \specialrule{.15em}{.05em}{.05em}
    Method & DiffMod & COCO & BSDS & KINS & FBowl & SV & Avg\\ 
    \hline
    \rowcolor{mygray}
    pix2gestalt &  SD2 &  {\color{blue}82.9} & \textbf{80.8} & 39.2 & {\color{blue}73.3} & 52.3 & 65.7\\
    Amodal Wild & - & \textbf{90.2} & - & - & - & - & - \\
    \hline
    SAM & - & 66.6 & 65.3 & 40.8 & 68.3 & 55.9 & 59.4\\
    SAM2 & - & 70.1 & 63.1 & 46.9 & 65.5 & 57.0 & 60.5\\
    Inpaint & SDXL & 76.5 & 74.2 & -    & -    & -   & -\\
    Ours & SDXL & 82.7 & {\color{blue}75.6} & \textbf{60.4} & 73.0 & 63.5 & \textbf{71.0}\\
    Ours & SD1.5 & 79.9 & 75.2 & 58.4 & 71.0 & \textbf{66.6} & 70.2\\
    Ours & SD2 & 73.2 & 72.6 & 57.2 & 72.8 & 57.2 & 66.6\\
    Ours &Flux & 75.5 & 75.5 & {\color{blue}60.2} & \textbf{75.2} & {\color{blue}65.6} & {\color{blue}70.4}\\
    \specialrule{.15em}{.05em}{.05em}
    \end{tabular}}
        \vspace{-0.2cm}
    \caption{Quantiative comparisons of amodal mask in mIoU(\%)\(\uparrow\). Methods except for pix2gestalt are tuning-free.
    The best result is bolded, and the second best is colored in {\color{blue}blue}.
    }
    \vspace{-0.4cm}
    \label{tab:miou}
\end{table}

\section{Experiments}
\label{sec:experiments}
Our method is tuning-free, and also a zero-shot amodal segmentation method. For a fair comparison, we strictly follow the experiment setup of zero-shot amodal segmentation experiment setting by~\citet{ozguroglu2024pix2gestalt} on COCO-A~\cite{cocoa} and BSDS-A~\cite{bsdsa}. To study the zero-shot capability, we evaluate three additional datasets, including KINS~\cite{kins}, FishBowl~\cite{fishbowl}, and SAILVOS~\cite{SAILVOS}. We report quantitative and qualitative results followed by ablations.

\subsection{Experiment setup}
We report on the following five datasets covering both real-world and synthetic images:\\
\ding{182} COCO-A~\cite{cocoa}: Based on COCO~\cite{lin2014microsoft} dataset, COCO-A~\cite{cocoa} is a human-annotated amodal segmentation dataset over natural images. We report on its evaluation set with 13k ground truth object amodal annotations in 2.5k images, including common objects.\\  
\ding{183} BSDS-A~\cite{bsdsa}: Derived from the Berkeley Segmentation Dataset (BSDS)~\cite{bsds}, BSDS-A~\cite{bsdsa} is an amodal segmentation dataset labeled with manual amodal annotation. We report on the evaluation image sets with 200 images from the real world.\\
\ding{184} KINS~\cite{kins}: KINS~\cite{kins}, derived from KITTI~\cite{kins_original} for autonomous driving, features manually annotated amodal masks and an evaluation set of 7k images.\\
\ding{185} FishBowl (Fbowl)~\cite{fishbowl} is a synthetic dataset that has different numbers of fish from an WebGL demo~\cite{webglsamplesWebGLAquarium}. Its evaluation set contains 1k videos of 128 frames each, with each frame treated independently for amodal segmentation.\\
\ding{186} SAILVOS (SV)~\cite{SAILVOS} is a synthetic dataset from the photo-realistic game GTA-V. It contains %
26k images along with 507k objects in the evaluation set.

{\bf\noindent Evaluation metric.}
Following~\citet{ozguroglu2024pix2gestalt}, we report the mean intersection over union (mIoU)
to evaluate predicted amodal masks. A higher mIoU indicates a better match of the prediction with the ground truth. We also report the mIoU over different subsets of the data based on the occlusion rate of the object. Specifically, we report on occlusion rate subsets that are less than 50\%. 
We observed that highly occluded objects yield uncertain annotations, as images often lack enough details for a complete amodal mask.

\begin{table}[t]
    \setlength{\tabcolsep}{3pt}
    \centering
    \resizebox{\linewidth}{!}{%
    \begin{tabular}{cccccccc}
    \specialrule{.15em}{.05em}{.05em}
    Method  & DiffMod & $\leq$50\% & $\leq$40\% & $\leq$30\% & $\leq$20\% & $\leq$10\% & $\leq$5\%\\ %
    \hline
    \rowcolor{mygray}
    pix2gestalt & SD2 & {\color{blue}83.1} & 83.7 & 84.3 & 85.2 & 86.7 & 87.0\\
    Amodal Wild & - & \textbf{86.7} & \textbf{88.3} & \textbf{88.6} & {\color{blue}89.9} & {\color{blue}93.3} & 92.2 \\
    \hline
    SAM  & -          & 68.4 & 71.2 & 73.7 & 76.3 & 79.9 & 81.2\\
    SAM2 & -          & 70.0 & 72.8 & 75.6 & 78.5 & 81.9 & 83.4\\
    Ours & SD1.5      & 79.9 & 82.9 & 85.9 & 88.8 & 92.1 & {\color{blue}93.6}\\
    Ours & SDXL       & 82.7 & {\color{blue}85.4} & {\color{blue}88.0} & \textbf{90.6} & \textbf{93.6} & \textbf{95.0} \\
    Ours & SD2       & 77.5 & 79.8 & 82.1 & 84.1 & 87.1  & 88.3\\
    Ours & Flux       & 76.2 & 77.2 & 78.7 & 80.4 & 83.2  & 84.1\\

    \specialrule{.15em}{.05em}{.05em}
    \end{tabular}
    }
        \vspace{-0.2cm}
    \caption{
    We compare the quality of amodal mask in mIoU(\%)$\uparrow$ from COCO-A~\cite{cocoa}.
    }
    \vspace{-0.5cm}
    \label{tab:cocoa}
\end{table}

{\bf \noindent Baselines.} We consider the state-of-the-art baseline of %
pix2gestalt, two training-required methods, and three additional tuning-free methods.\\
\ding{192} pix2gestalt~\cite{ozguroglu2024pix2gestalt} takes an RGB image and its modal mask to generate an amodal mask by using SAM~\cite{sam} to collect on a customized training dataset that has more than 800k image pairs with occlusions to a fine-tuned %
a \textit{pre-trained diffusion model} of StableDiffusion2 (SD2)~\cite{SD2}.\\
\ding{193} Amodal Wild~\cite{zhan2024amodal} uses a two-stage approach. First, an occluder mask is predicted from an RGB image and its modal mask. Next, a U-Net-based model leveraging features from a pre-trained Stable Diffusion (using the modal mask and occluder boundary) predicts the amodal mask.\\
\ding{194} Inpaint-SDXL~\cite{podell2023sdxl}: Given a visible mask and an RGB image, SDXL inpaints its region by leveraging a pre-trained model to remove missing pixel information. This baseline is proposed by~\citet{ozguroglu2024pix2gestalt} in pix2gestalt. We directly report the number from their paper, as the code for this baseline has not been released. 
\\
\ding{195} SAM~\cite{sam}: takes a set of points and an RGB image to segment pixels that fall into the same object category based on features from an image encoder. This is a strong modal baseline, as reported by~\citet{ozguroglu2024pix2gestalt}\\
\ding{196} SAM2~\cite{sam2}: We also consider a even strong modal baseline of SAM2, which is a improved version of SAM.

We also tried comparing to Amodal Completion by~\citet{progressive}, however, as it is limited to 83 categories, the approach was unable to generate a prediction for many images in the datasets we considered.

{\bf \noindent Implementation.} %
We consider several popular diffusion models, including Stable Diffusion 1.5 and 2 (SD1.5, SD2)~\cite{Rombach_2022_CVPR}, Stable Diffusion XL (SDXL)~\cite{podell2023sdxl}, and Flux~\cite{flux}. 
We set $\mathfrak{s}=0.3$, $w=7.5$ for Stable Diffusion 1.5~\cite{Rombach_2022_CVPR} and Stable Diffusion XL~\cite{podell2023sdxl}, and $w=1.5$ for Flux~\cite{flux}. Images are refined with 20 iterative steps for amodal completion, and the mask $\mM$ is extracted by uniformly sampling nine points from $\mV$ using SAM~\cite{sam}.  All experiments were performed on an NVIDIA RTX 4090 (24GB VRAM), and 8-bit quantization was applied for Flux to reduce the memory usage.

\begin{table}[t]
    \setlength{\tabcolsep}{3pt}
    \centering
    \resizebox{\linewidth}{!}{%
    \begin{tabular}{cccccccc}
    \specialrule{.15em}{.05em}{.05em}
    Method & DiffMod & $\leq$50\% & $\leq$40\% & $\leq$30\% & $\leq$20\% & $\leq$10\% & $\leq$5\%\\ %
    \hline
    \rowcolor{mygray}
    pix2gestalt & SD2 & {\color{blue}77.1} & 78.0 & 79.0 & 80.2 & 81.4 & 81.9\\
    \hline
    SAM  & -          & 62.1 & 64.6 & 66.8 & 69.3 & 72.6 & 74.9\\
    SAM2 & -          & 63.4 & 66.6 & 69.4 & 72.4 & 76.7 & 78.6\\
    Ours & SD1.5      & \textbf{78.3} & \textbf{80.5} & {\color{blue}82.5} & 84.5 & 86.7 & 87.4\\
    Ours & SDXL       & 75.7 & {\color{blue}79.2} & 82.2 & {\color{blue}85.7} & \textbf{88.9} & \textbf{90.7}\\
    Ours & SD2        & 76.2 & 79.8 & \textbf{82.8} & \textbf{85.8} & {\color{blue}88.7} & {\color{blue}90.1}\\
    Ours & Flux       & 76.4 & 78.5 & 81.0 & 84.1 & 85.9 & 72.6\\

    \specialrule{.15em}{.05em}{.05em}
    \end{tabular}
    }
    \vspace{-0.2cm}
    \caption{We compare the quality of amodal mask in  mIoU(\%)$\uparrow$ from BSDS-A~\cite{bsdsa}.}
    \label{tab:bsds}
    \vspace{-0.3cm}
\end{table}

\subsection{Quantitative results}
{\noindent\bf Main results.}~\tabref{tab:miou} reports mIoU using five datasets, and we \textbf{bold} the best metric and {\color{blue}colored} the second best metric. 
For the COCO-A~\cite{cocoa} and BSDS-A~\cite{bsdsa}, %
pix2gestalt~\cite{ozguroglu2024pix2gestalt} (a non-tuning free method) has the best metrics, followed by Stable Diffusion XL with ours by 0.2\% and 5.2\%.

We highlight that the first two datasets are reported in %
pix2gestalt to study zero-shot amodal segmentation. Recall that their approach trains on a ``synthetically curated dataset'' and is hence zero-shot. However, it is unclear whether this curated dataset generalizes beyond COCO-A and BSDS-A~\eg, what if the testing distribution is very different from their curated data.

We report on three additional datasets with various object categories to further study the zero-shot capability. On KINS~\cite{kins}, FishBowl (FBowl)~\cite{fishbowl}, SAILVOS (SV)~\cite{SAILVOS}, show that our methods generate 21.2\%, 1.9\%, and 14.3\% more accurate mask than pix2gestalt. Importantly, our method performs best from the tuning-free approaches and convincingly outperforms the modal baseline. We could not compare on Inpainting-SDXL, as the code was not released, and we could not reproduce it.

Below, we report and discuss the detailed results based on different object occlusion rates. In the appendix, we provide further qualitative analysis for each dataset.

\begin{table}[t]
    \setlength{\tabcolsep}{3pt}
    \centering
    \resizebox{\linewidth}{!}{%
    \begin{tabular}{cccccccc}
    \specialrule{.15em}{.05em}{.05em}
    Method & DiffMod & $\leq$50\% & $\leq$40\% & $\leq$30\% & $\leq$20\% & $\leq$10\% & $\leq$5\%\\ %
    \hline
    \rowcolor{mygray}
    pix2gestalt & SD2 & 39.1 & 40.2 & 41.7 & 43.7  & 48.0 & 55.1\\
    \hline
    SAM  & -                          & 42.2 & 42.8 & 43.7 & 44.8 & 47.3 & 50.8\\
    SAM2 & -                    & 48.7 & 49.6 & 50.7 & 52.3 & 54.9 & 58.3\\
    Ours & SD1.5           & \textbf{64.8} & \textbf{66.2} & \textbf{68.2} & \textbf{70.6} & \textbf{74.1} & \textbf{77.3}\\
    Ours & SDXL               & 60.4 & 62.1 & 64.1 & 66.6 & 70.8 & 75.1\\
    Ours & SD2               &  60.2 & 61.4 & 63.0 & 64.8 & 68.1 & 73.5\\
    Ours & Flux                         & {\color{blue}64.7} & {\color{blue}65.4} & {\color{blue}67.3} & {\color{blue}69.5} & {\color{blue}72.7} & {\color{blue}75.6}\\

    \specialrule{.15em}{.05em}{.05em}
    \end{tabular}
    }
    \vspace{-0.1cm}
    \caption{We compare the quality of amodal mask in mIoU(\%)$\uparrow$ from KINS~\cite{kins}.}
    \label{tab:kins}
        \vspace{-0.2cm}
\end{table}

\begin{table}[t]
    \setlength{\tabcolsep}{3pt}
    \centering
    \resizebox{\linewidth}{!}{%
    \begin{tabular}{cccccccc}
    \specialrule{.15em}{.05em}{.05em}
    Method & DiffMod & $\leq$50\% & $\leq$40\% & $\leq$30\% & $\leq$20\% & $\leq$10\% & $\leq$5\%\\ %
    \hline
    \rowcolor{mygray}
    pix2gestalt & SD2 & 77.1 & 78.0 & 79.0 & 80.2 & 81.4 & 81.9\\
    \hline
    SAM & -                           & 60.9 & 63.0 & 65.0 & 67.3 & 70.4 & 74.3\\
    SAM2 & -                        & 68.0 & 69.8 & 72.4 & 74.4 & 77.1 & 79.8\\
    Ours & SD1.5          & 79.0 & 81.3  & {\color{blue}83.8} & {\color{blue}86.7} & {\color{blue}89.6} & {\color{blue}90.1}\\
    Ours & SDXL              & {\color{blue}79.5} & {\color{blue}81.1} & 83.6 & 85.3 & 87.1 & 87.5\\
    Ours & SD2               & 77.9 & 79.3 & 80.7 & 81.9 & 82.7 & 82.9\\
    Ours & Flux                         & \textbf{80.4} & \textbf{82.8} & \textbf{85.3} & \textbf{88.0} & \textbf{90.7} & \textbf{91.9}\\
    \specialrule{.15em}{.05em}{.05em}
    \end{tabular}
    }
        \vspace{-0.2cm}
    \caption{We compare the quality of amodal mask in mIoU(\%)$\uparrow$ from FishBowl~\cite{fishbowl}.}
    \label{tab:fish}
        \vspace{-0.55cm}
\end{table}

\begin{table}[t]
    \setlength{\tabcolsep}{2pt}
    \centering
    \resizebox{\linewidth}{!}{%
    \begin{tabular}{cccccccc}
    \specialrule{.15em}{.05em}{.05em}
    Method & DiffMod & $\leq$50\% & $\leq$40\% & $\leq$30\% & $\leq$20\% & $\leq$10\% & $\leq$5\%\\ %
    \hline
    \rowcolor{mygray}
    pix2gestalt & SD2 & 59.4 & 60.2 & 61.3 & 62.9 & 65.6 & 67.8\\
    \hline
    SAM & -                           & 66.2 & 68.7 & 71.1 & 73.5 & 76.2 & 79.3\\
    SAM2 & -                         & 67.5 & 69.9 & 72.2 & 74.5 & 76.8 & 79.2\\
    Ours & SD1.5        & \textbf{80.0} & \textbf{82.3} & \textbf{84.1} & \textbf{85.8} & \textbf{87.4} & {\color{blue}88.7}\\
    Ours & SDXL               & 62.3 & 64.0 & 65.7 & 67.5 & 70.5 & 72.1\\
    Ours & SD2               & 68.6 & 70.1 & 71.6 & 73.2 & 75.9 & 77.7\\
    Ours & Flux                         & {\color{blue}79.3} & {\color{blue}81.8} & {\color{blue}83.6} & {\color{blue}85.4} & {\color{blue}87.2} & \textbf{88.8}\\
    \specialrule{.15em}{.05em}{.05em}
    \end{tabular}
    }
    \vspace{-0.1cm}
    \caption{%
    We compare the quality of amodal mask in mIoU(\%)$\uparrow$ from SAILVOS.}
    \label{tab:sailvos}
    \vspace{-0.5cm}
\end{table}

{\noindent\bf Detailed COCO-A results.}
\cref{tab:cocoa} shows the performance based on our approach using four foundation models (Stable Diffusion 1.5~\cite{Rombach_2022_CVPR}, Stable Diffusion XL~\cite{podell2023sdxl}, Stable Diffusion 2~\cite{SD2}, Flux~\cite{flux}) to pix2gestalt~\cite{ozguroglu2024pix2gestalt} along with SAM~\cite{sam} and SAM2~\cite{sam2}.

Our work using Stable Diffusion XL~\cite{podell2023sdxl} generates 1.7\%, 3.7\%, 5.4\%, 6.9\%, and 8.0\%  more accurate amodal masks on average when objects are occluded less than equal to 40\%, 30\%, 20\%, 10\%, and 5\% compared to pix2gestalt, respectively and it has a 0.2\% higher mIoU as shown~\cref{tab:miou}.

While the training-required method~\cite{zhan2024amodal} achieves a 7.\% mIoU gain, our approach improves 0.7\%, 0.3\%, and 2.8\% at 20\%, 10\%, and 5\% occlusion, respectively.

\begin{figure*}[ht]
    \centering
    \small
    \setlength{\tabcolsep}{0.5pt} %
    \renewcommand{\arraystretch}{0.05} %

    \begin{tabular}{cccccccc} %
        \multicolumn{1}{c}{Image} &
        \multicolumn{1}{c}{Visible Mask} &
        \multicolumn{1}{c}{Ours (SD1.5)} &
        \multicolumn{1}{c}{Ours (SDXL)} &
        \multicolumn{1}{c}{Ours (SD2)} &
        \multicolumn{1}{c}{Ours (Flux)} &
        \multicolumn{1}{c}{pix2gestalt} &
        \multicolumn{1}{c}{GT} \\

        \includegraphics[width=0.12\linewidth]{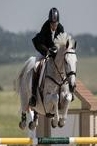} &
        \includegraphics[width=0.12\linewidth]{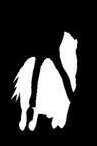} &
        \includegraphics[width=0.12\linewidth]{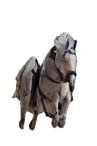} &
        \includegraphics[width=0.12\linewidth]{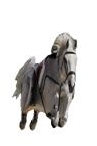} &
        \includegraphics[width=0.12\linewidth]{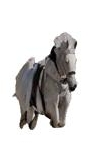} &
        \includegraphics[width=0.12\linewidth]{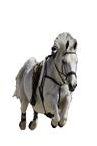} &
        \includegraphics[width=0.12\linewidth]{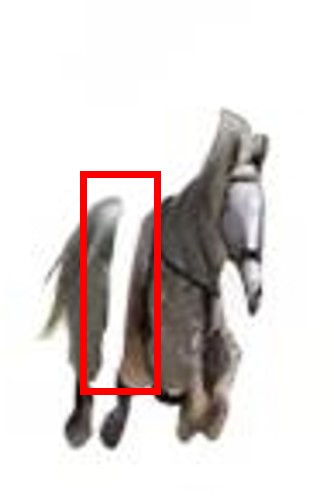} &
        \includegraphics[width=0.12\linewidth]{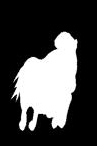} \\

        \includegraphics[width=0.12\linewidth]{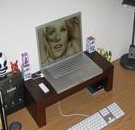} &
        \includegraphics[width=0.12\linewidth]{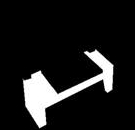} &
        \includegraphics[width=0.12\linewidth]{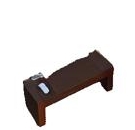} &
        \includegraphics[width=0.12\linewidth]{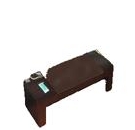} &
        \includegraphics[width=0.12\linewidth]{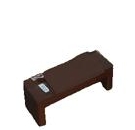} &
        \includegraphics[width=0.12\linewidth]{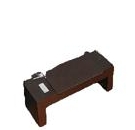} &
        \includegraphics[width=0.12\linewidth]{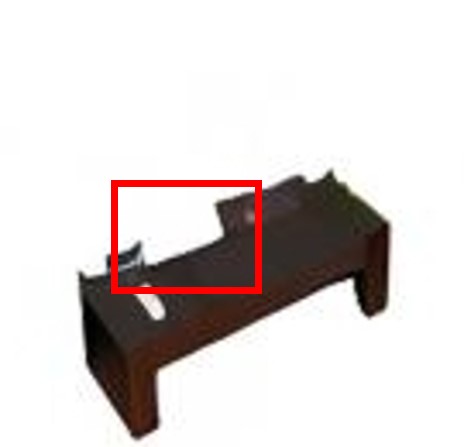} &
        \includegraphics[width=0.12\linewidth]{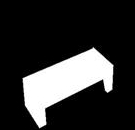} \\

        \includegraphics[width=0.12\linewidth]{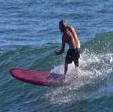} &
        \includegraphics[width=0.12\linewidth]{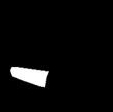} &
        \includegraphics[width=0.12\linewidth]{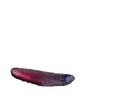} &
        \includegraphics[width=0.12\linewidth]{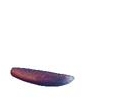} &
        \includegraphics[width=0.12\linewidth]{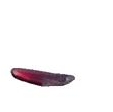} &
        \includegraphics[width=0.12\linewidth]{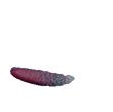} &
        \includegraphics[width=0.12\linewidth]{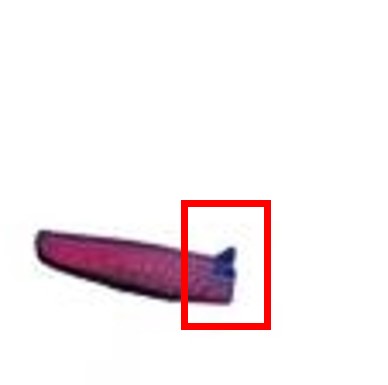} &
        \includegraphics[width=0.12\linewidth]{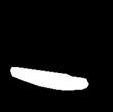} \\

        \includegraphics[width=0.12\linewidth]{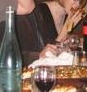} &
        \includegraphics[width=0.12\linewidth]{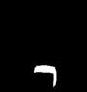} &
        \includegraphics[width=0.12\linewidth]{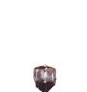} &
        \includegraphics[width=0.12\linewidth]{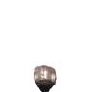} &
        \includegraphics[width=0.12\linewidth]{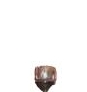} &
        \includegraphics[width=0.12\linewidth]{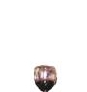} &
        \includegraphics[width=0.12\linewidth]{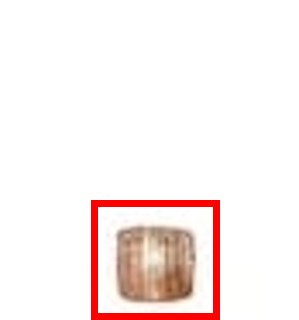} &
        \includegraphics[width=0.12\linewidth]{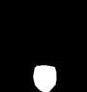} \\

    \end{tabular}
        \vspace{-0.15cm}
    \caption{We compare the accuracy of amodal mask in COCO-A~\cite{cocoa} and BSDS-A~\cite{bsdsa} using various diffusion-based inpainting models. Keep in mind that we focus on generating accurate amodal masks rather than on synthesizing an accurate and high-quality image. We highlight incomplete and out-of-shape areas using a \textcolor{red}{red box.}
    }
    \label{fig:vis}
        \vspace{-0.1cm}
\end{figure*}

{\noindent\bf Detailed BSDS-A results.}
In~\tabref{tab:bsds}, our method hash higher 1.2\% and 2.5\% mIoU with Stable Diffusion 1.5~\cite{Rombach_2022_CVPR} at 50\% and 40\% occlusion, 3.8\% and 5.6\% with Stable Diffusion 2~\cite{SD2} at 30\% and 20\%, and 5.9\% and 6.6\% with Stable Diffusion XL~\cite{podell2023sdxl} at 10\% and 5\%, compared to pix2gestalt~\cite{ozguroglu2024pix2gestalt}. Our method achieves 4.9\% more accurate amodal mask on average for occlusions under 50\%.\\
{\noindent\bf Detailed KINS results.}
In~\tabref{tab:kins}, Stable Diffusion 1.5~\cite{Rombach_2022_CVPR} with our approach predicts 25.7\%, 26\%, 26.5\%, 26.9\%, 26.1\%, and 22.2\% more accurate amodal masks, in all occlusion rates, compared to pix2gestalt~\cite{ozguroglu2024pix2gestalt}. On average, our method outperforms 25.6\% to generate an accurate amodal mask with occlusion rates of 50\% or less. 
The largest performance gap among the five datasets comes from pix2gestalt's training data, which is insufficient to cover KINS~\cite{kins} occlusions, leading to robustness issues.
\\
{\noindent\bf Detailed FishBowl results.}
In~\cref{tab:fish}, our approach with Flux~\cite{flux} shows 1.2\%, 3.3\%, 5.7\%, 8.4\%, and 9.6\% more accurate amodal mask than pix2gestalt~\cite{ozguroglu2024pix2gestalt}. For occlusion rates of 50\% or less, our method predicts an average mIoU that is 6.9\% higher than pix2gestalt, further demonstrating the zero-shot capabilities.\\
{\noindent\bf Detailed SAILVOS results.}
\cref{tab:sailvos} shows that our approach generates 20.6\%, 22.1\%, 22.8\%, 22.9\%, 21.8\%, 21\% more precise amodal mask than pix2gestalt~\cite{ozguroglu2024pix2gestalt}, averaging 21.9\% higher mIoU for occlusions $\leq$ 50\%.

\begin{figure*}[t]
    \centering
    \small
    \setlength{\tabcolsep}{0.2pt} %
    \renewcommand{\arraystretch}{0.3} %

    \begin{tabular}{ccccccccc} %
        &
        \multicolumn{1}{c}{Input} &
        \multicolumn{1}{c}{Visible Mask}  &
        \multicolumn{1}{c}{Ours (SD1.5)} &
        \multicolumn{1}{c}{Ours (SDXL)} &
        \multicolumn{1}{c}{Ours (SD2)} &
        \multicolumn{1}{c}{Ours (Flux)} &
        \multicolumn{1}{c}{pix2gestalt} &
        \multicolumn{1}{c}{GT} \\

        \raisebox{1.\height}{\rotatebox{90}{KINS}} & \includegraphics[width=0.12\linewidth]{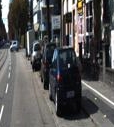} &
        \includegraphics[width=0.12\linewidth]{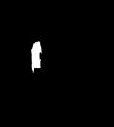} &
        \includegraphics[width=0.12\linewidth]{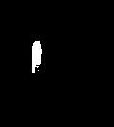} &
        \includegraphics[width=0.12\linewidth]{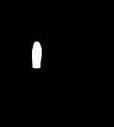} &
        \includegraphics[width=0.12\linewidth]{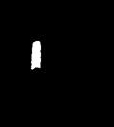} &
        \includegraphics[width=0.12\linewidth]{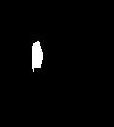} &
        \includegraphics[width=0.12\linewidth]{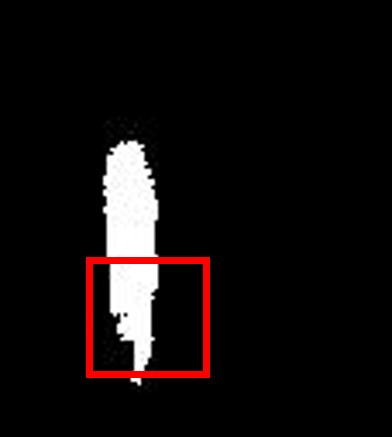} &
        \includegraphics[width=0.12\linewidth]{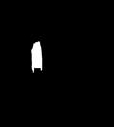} \\

        \raisebox{.35\height}{\rotatebox{90}{FishBowl}} & \includegraphics[width=0.12\linewidth]{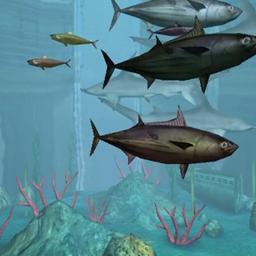} &
        \includegraphics[width=0.12\linewidth]{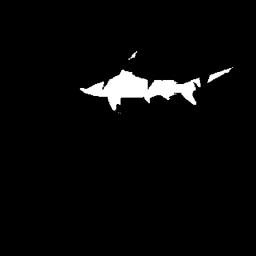} &
        \includegraphics[width=0.12\linewidth]{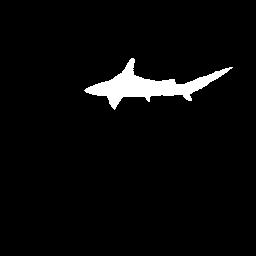} &
        \includegraphics[width=0.12\linewidth]{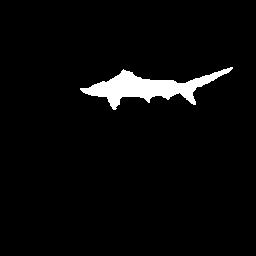} &
        \includegraphics[width=0.12\linewidth]{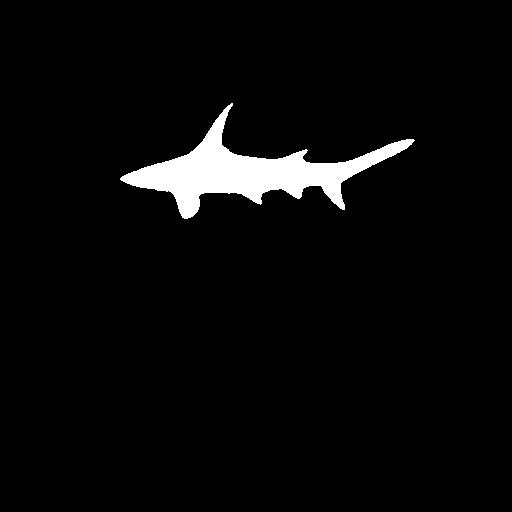} &
        \includegraphics[width=0.12\linewidth]{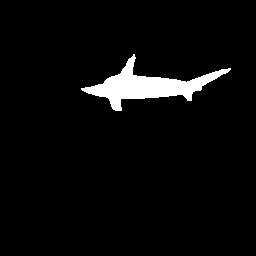} &
        \includegraphics[width=0.12\linewidth]{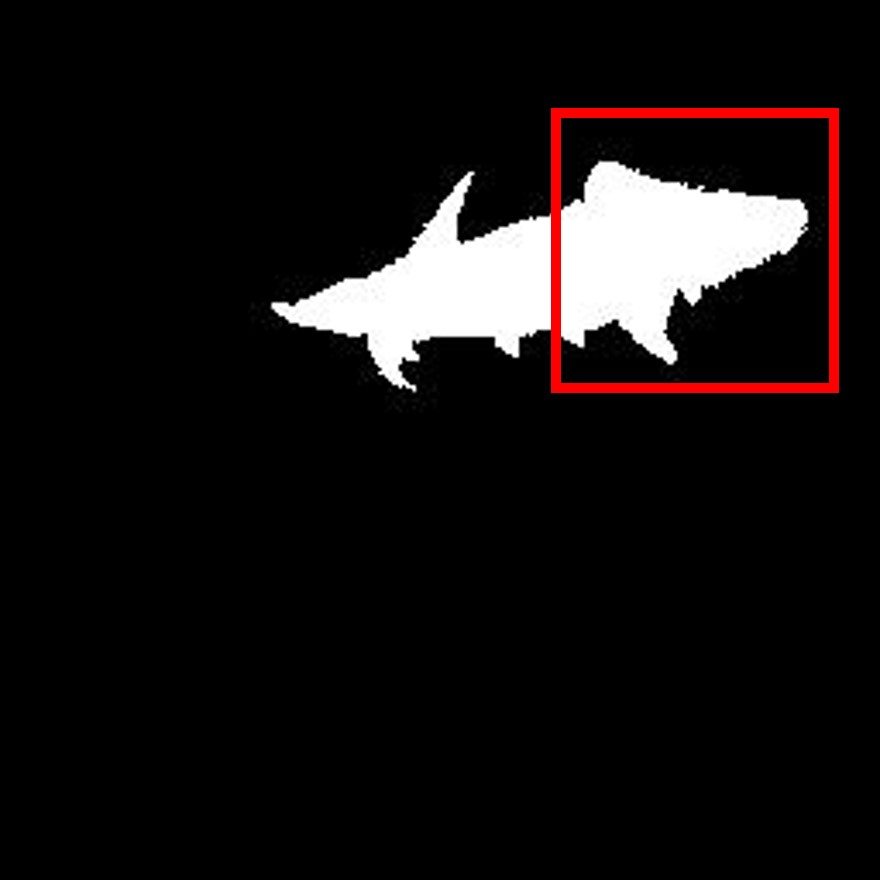} &
        \includegraphics[width=0.12\linewidth]{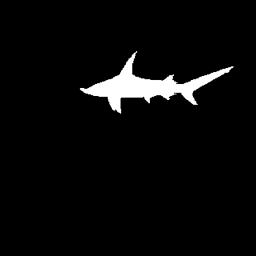} \\

        \raisebox{.3\height}{\rotatebox{90}{SAILVOS}} & \includegraphics[width=0.12\linewidth]{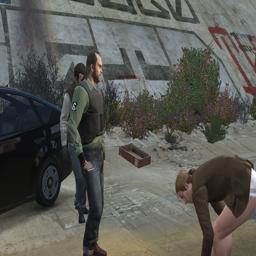} &
        \includegraphics[width=0.12\linewidth]{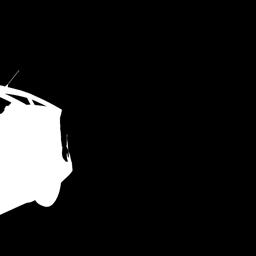} &
        \includegraphics[width=0.12\linewidth]{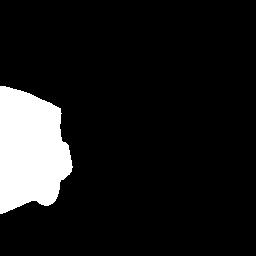} &
        \includegraphics[width=0.12\linewidth]{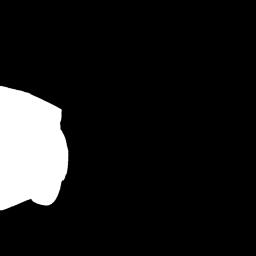} &
        \includegraphics[width=0.12\linewidth]{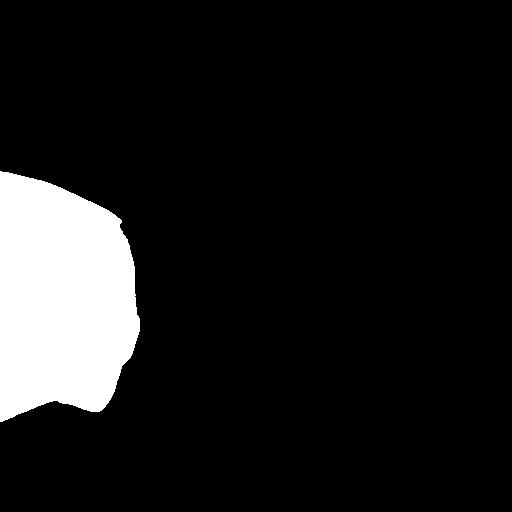} &
        \includegraphics[width=0.12\linewidth]{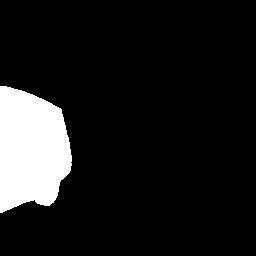} &
        \includegraphics[width=0.12\linewidth]{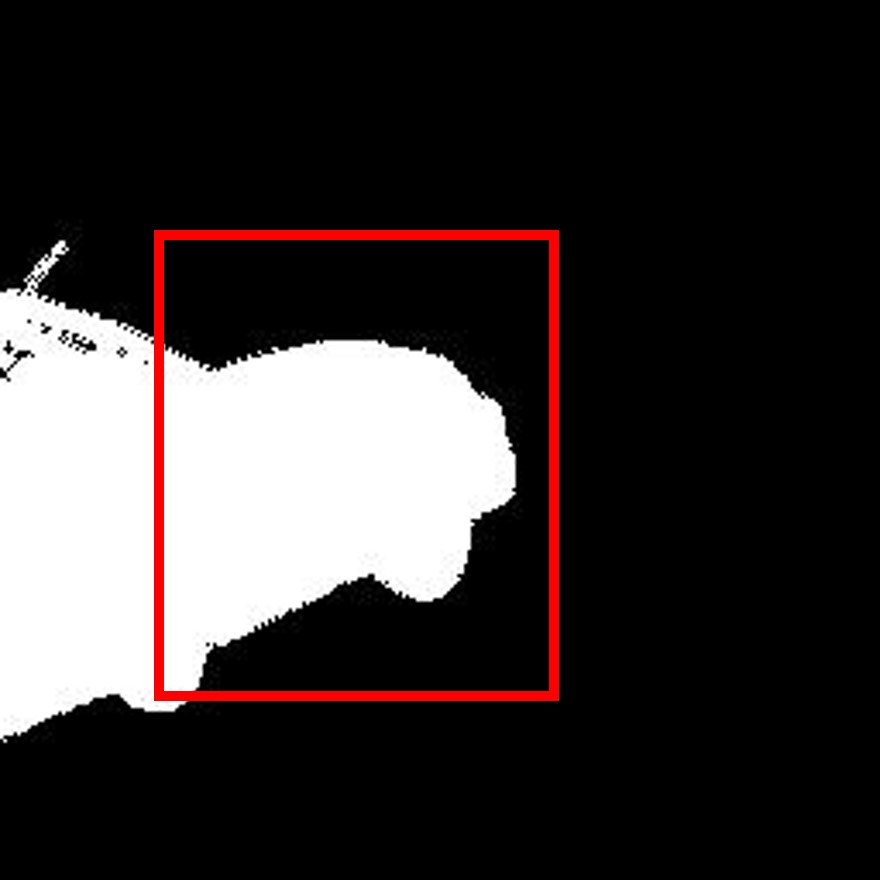} &
        \includegraphics[width=0.12\linewidth]{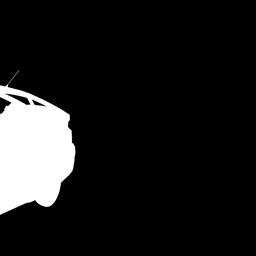} \\

    \end{tabular}
        \vspace{-0.1cm}
    \caption{Qualitative comparison of amodal mask on  KINS~\cite{kins}, FishBowl~\cite{fishbowl}, and SAILVOS~\cite{SAILVOS}. We observe that for novel categories/~domain pix2gestalt may hallucinate inaccurate amodal masks.
    }
       \vspace{-0.35cm} 
    \label{fig:kins_fish_sailvos}
\end{figure*}

\subsection{Qualitative results}
We compare the visual quality of the amodal mask in~\figref{fig:vis}. Please remember that our goal is not to generate high-quality inpainted results. That is, one should judge how closely the contour of the inpainted object matches the ground truth. Our method successfully predicts the occluded regions, \eg, removing the occlusion and generating a complete horse. On the other hand, pix2gestalt fails to generate in the occluded areas (rows 1 and 2) and struggles to handle the outside of the visible area (rows 3 and 4).

Next, we further show the amodal mask generation results on the three additional datasets, including KINS~\cite{kins} (row 1), FishBowl~\cite{fishbowl} (row 2), and SAILVOS~\cite{SAILVOS} (row 3) in~\cref{fig:kins_fish_sailvos}. The first row shows that pix2gestalt overextends the car. The second row demonstrates that pix2gestalt misunderstands the visual context and adds ``hallucinations'', which is another car, to generate amodal masks.
Overall, we observe that pix2gestalt performs worse on these additional datasets, possibly due to a larger gap in distribution from their curated data. In contrast, our method shows robustness with high-quality amodal masks.

\begin{figure}[t]
    \centering
    \small
    \setlength{\tabcolsep}{0.5pt} %
    \renewcommand{\arraystretch}{0.1} %

    \begin{tabular}{cccc} %
        \multicolumn{1}{c}{Image} &
        \multicolumn{1}{c}{Mask} &
        \multicolumn{1}{c}{Ours} &
        \multicolumn{1}{c}{pix2gestalt} \\

        \subfloat{\includegraphics[width=0.25\linewidth]{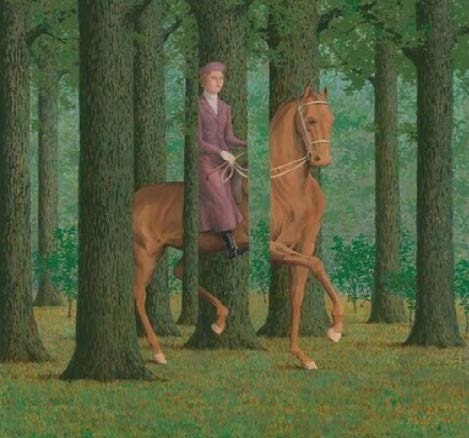}} &
        \subfloat{\includegraphics[width=0.25\linewidth]{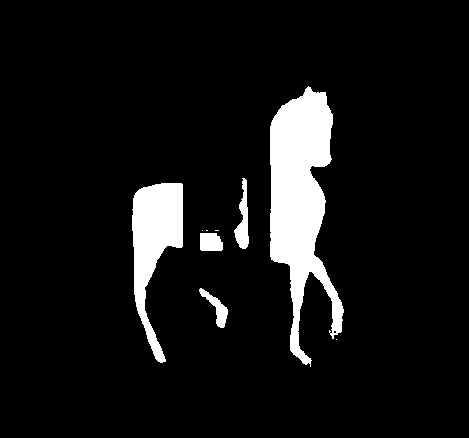}} &
        \subfloat{\includegraphics[width=0.25\linewidth]{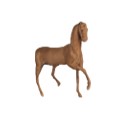}} &
        \subfloat{\includegraphics[width=0.25\linewidth]{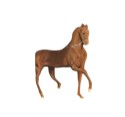}} \\

        \subfloat{\includegraphics[width=0.25\linewidth]{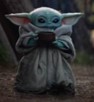}} &
        \subfloat{\includegraphics[width=0.25\linewidth]{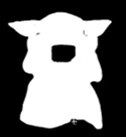}} &
        \subfloat{\includegraphics[width=0.25\linewidth]{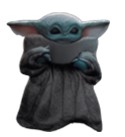}} &
        \subfloat{\includegraphics[width=0.25\linewidth]{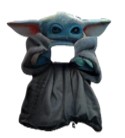}} \\

    \end{tabular}
        \vspace{-0.2cm}
    \caption{Amodal completion results on in-the-wild images comparing Ours (SDXL) and pix2gestalt~\cite{ozguroglu2024pix2gestalt}.
    }
    \label{fig:compare2pix2gestalt}
\end{figure}

\begin{figure}[t]
    \centering
    \setlength{\tabcolsep}{0.1pt} %
    \renewcommand{\arraystretch}{0.5} %

    \begin{tabular}{ccccc} %
        \multicolumn{1}{c}{GT} &
        \multicolumn{1}{c}{Ours} &
        \multicolumn{1}{c}{Mask} &
        \multicolumn{1}{c}{Diff} &
        \multicolumn{1}{c}{Ours RGB} \\

        \includegraphics[width=0.2\linewidth]{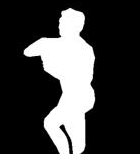} &
        \includegraphics[width=0.2\linewidth]{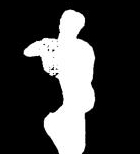} &
        \includegraphics[width=0.2\linewidth]{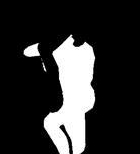} &
        \includegraphics[width=0.2\linewidth]{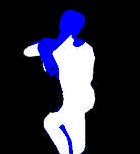} &
        \includegraphics[width=0.2\linewidth]{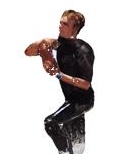}\\

        \includegraphics[width=0.2\linewidth]{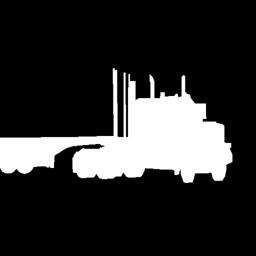} &
        \includegraphics[width=0.2\linewidth]{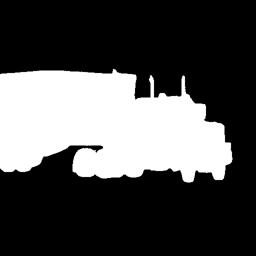} &
        \includegraphics[width=0.2\linewidth]{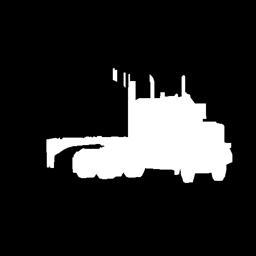} &
        \includegraphics[width=0.2\linewidth]{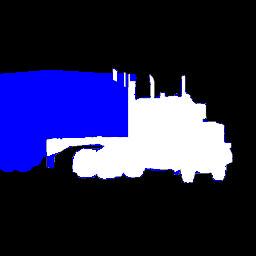} &
        \includegraphics[width=0.2\linewidth]{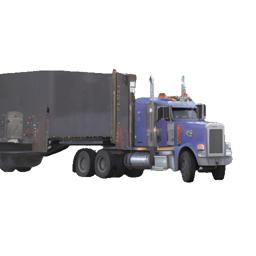}\\

        \includegraphics[width=0.2\linewidth]{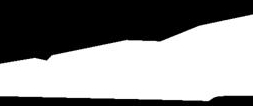} &
        \includegraphics[width=0.2\linewidth]{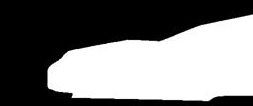} &
        \includegraphics[width=0.2\linewidth]{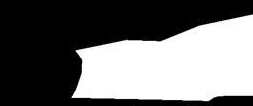} &
        \includegraphics[width=0.2\linewidth]{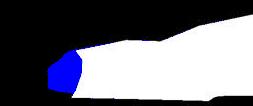} &
        \includegraphics[width=0.2\linewidth]{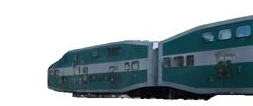}\\

    \end{tabular}
        \vspace{-0.25cm}
    \caption{\textcolor{blue}{Blue} pixels denote differences between the visible mask and our predicted mask.
    }
    \label{fig:extrapolation}
        \vspace{-0.3cm}
\end{figure}

We also experimented with in-the-wild images to validate our approaches compared to pix2gestalt (\figref{fig:compare2pix2gestalt}). We start with the horse (first row) that pix2gestalt reported in their paper. The predicted mask from our approach shows a comparable quality to that generated by pix2gestalt. We also show another example (second row), where ours completes a cloth behind the cup that Grogu is holding, while pix2gestalt barely made any changes to the input image. %
Moreover, \cref{fig:extrapolation} shows cases of amodal mask extrapolation, demonstrating our method's robustness.

\subsection{Ablation studies \& analysis}
{\bf\noindent Ablations.} We evaluate the effectiveness of each component by removing them in~\cref{tab:ab_components}. The experiment is conducted on COCO-A~\cite{cocoa} using Stable Diffusion XL~\cite{podell2023sdxl}. The first row shows the mIoU with all the components included. When the leaking conditioning (\cref{sec:leakage_cond}), the context-aware background (\cref{sec:cond_img}), or the inpainting area (\cref{sec:inpaint_area}) is excluded, the accuracy of the amodal mask falls by 37.6\%, 5.6\%, 6.2\% in the mIoU compared to using all components respectively and the greater drop indicates higher importance for generating accurate amodal masks.
Moreover, we ablate the value of $\mathfrak{s}$ (\cref{tab:rebuttal_s_effect_cocoa}). In~\cref{tab:rebut_ablation}, we verify the effectiveness of our designed inpainting region, $\mM$, along with the context-aware background by comparing it to a simple rectangular mask with a white background.

{\bf\noindent Computation efficiency.} We compare the efficiency of our approach to pix2gestalt. For the smallest model, SD2 is 4.1$\times$ more efficient in memory, and the inference (0.3 seconds) is 19$\times$ faster compared to pix2gestalt~\cite{ozguroglu2024pix2gestalt}. Additionally, our best model (SDXL) is also more efficient than pix2gestalt. See \tabref{tab:run_stats} in the Appendix for more results.

\begin{table}[t]
    \small
    \centering
    \resizebox{\linewidth}{!}{%
    \begin{tabular}{ccccc}
    \specialrule{.15em}{.05em}{.05em}
    DiffMod & Leakage & Background & Mask & mIoU(\%)$\uparrow$ \\
    SDXL & \ding{51} & \ding{51} & \ding{51} & 76.5 \\
    \hline
    SDXL & {\color{red}\ding{55}} & \ding{51} & \ding{51} & 38.9 \\
    SDXL & \ding{51} & {\color{red}\ding{55}}  & \ding{51} & 70.9 \\
    SDXL & \ding{51} & \ding{51} & {\color{red}\ding{55}}  & 70.3 \\
    \specialrule{.15em}{.05em}{.05em}
    \end{tabular}
    }
        \vspace{-0.25cm}
    \caption{We show our component's effectiveness using mIoU by excluding each component from the pipeline.}
    \vspace{-0.15cm}
    \label{tab:ab_components}
\end{table}

\begin{table}[t]
    \small
    \centering
    \resizebox{0.99\linewidth}{!}{%
    \begin{tabular}{ccccccc}
    \specialrule{.15em}{.05em}{.05em}
    $\mathfrak{s}$ & $\leq$50\% & $\leq$40\% & $\leq$30\% & $\leq$20\% & $\leq$10\% & $\leq$5\%\\ %
    \midrule
    0.1  & 75.0 & 77.5 & 80.1 & 82.5 & 86.3 & 87.9 \\
    0.15 & {\color{blue}75.8} & {\color{blue}78.2} & {\color{blue}80.7} & {\color{blue}83.0} & {\color{blue}86.6} & {\color{blue}88.1} \\
    \rowcolor{mygray}
    0.3 & \textbf{79.9} & \textbf{82.9} & \textbf{85.9} & \textbf{88.8} & \textbf{92.1} & \textbf{93.6} \\
    0.45 & 75.1 & 77.2 & 79.2 & 81.2 & 84.4 & 85.5 \\
    0.6 & 69.6 & 71.3 & 73.1 & 75.1 & 78.5 & 79.4 \\
    0.9 & 60.1 & 61.6 & 63.0 & 64.4 & 66.1 & 67.1 \\
    \specialrule{.15em}{.05em}{.05em}
    \end{tabular}
    }
    \vspace{-0.22cm}
    \caption{The impact of $\mathfrak{s}$ using mIoU(\%) from COCO-A.}
    \vspace{-0.2cm}
    \label{tab:rebuttal_s_effect_cocoa}
\end{table}

\begin{table}[t]
    \setlength{\tabcolsep}{3pt}
    \centering
    \resizebox{0.99\linewidth}{!}{%
    \begin{tabular}{cccccccc}
    \specialrule{.15em}{.05em}{.05em}
    $\mM$-type & background & $\leq$50\% & $\leq$40\% & $\leq$30\% & $\leq$20\% & $\leq$10\% & $\leq$5\%\\ %
    \midrule
    \rowcolor{mygray}
    Ours & Ours & \textbf{79.9} & \textbf{82.9} & \textbf{85.9} & \textbf{88.8} & \textbf{92.1} & \textbf{93.6} \\
    Rect. & Ours & {\color{blue}77.7} & {\color{blue}80.0} & {\color{blue}82.1} & {\color{blue}84.1} & {\color{blue}86.9} & {\color{blue}88.2} \\
    Ours & White & 71.4 & 73.8 & 76.2 & 78.6 & 81.8 & 83.1 \\
    Rect. & White & 70.3 & 72.9 & 75.2 & 77.7 & 81.0 & 82.3 \\

    \specialrule{.15em}{.05em}{.05em}
    \end{tabular}
    }
    \vspace{-0.2cm}
    \caption{Comparisons with a different mask and color histogram.}
    \vspace{-0.25cm}
    \label{tab:rebut_ablation}
\end{table}

 \section{Conclusion}
We introduce a tuning-free/zero-shot amodal segmentation method by leveraging the occlusion-fere bias of pre-trained diffusion inpainting models. Our approach customizes the conditioning image, designs a new inpainting region, and uses a novel leakage conditioning technique. Experiments on five datasets demonstrate that our model (SDXL) improves mIoU by \textbf{5.3\%}, with \textbf{4.8$\times$} faster inference and \textbf{1.4$\times$} VRAM efficiency over pix2gestalt. Other models (SD1.5, SD2, Flux) are also effective. As diffusion inpainting techniques continue to improve, we anticipate further advancements in segmentation performance.

\clearpage
{
    \small
    \bibliographystyle{ieeenat_fullname}
    \bibliography{reference}
}
\clearpage
\clearpage

\maketitlesupplementary
\setcounter{figure}{0}
\setcounter{section}{0}
\setcounter{table}{0}
\renewcommand\thefigure{S\arabic{figure}}
\renewcommand\thesection{S\arabic{section}}
\renewcommand\thetable{S\arabic{table}}
\renewcommand{\lstlistingname}{Code} %
\setcounter{page}{1}

{\bf \noindent The appendix is organized as follows:}
\begin{itemize}
    \item In~\cref{sec:supp_qa}, we provide additional qualitative results.
    \item In~\cref{sec:supp_ab}, we conduct additional ablations by comparing two different ways of creating mask regions.
    \item In~\cref{sec:supp_implementation}, we provide implementation details.
\end{itemize}

\section{Additional Qualitative Analysis}
\label{sec:supp_qa}
We demonstrate a more detailed qualitative analysis using the four foundation diffusion-based inpainting models, SD1.5~\cite{Rombach_2022_CVPR}, SD2~\cite{SD2}, SDXL~\cite{podell2023sdxl}, Flux~\cite{flux}, and pix2gestalt~\cite{ozguroglu2024pix2gestalt}, which is the SOTA, from the five diverse image datasets. We put a red bounding box around a defective part for visualization purposes.

\begin{figure}[ht]
    \centering
    \includegraphics[width=1.0\linewidth]
    {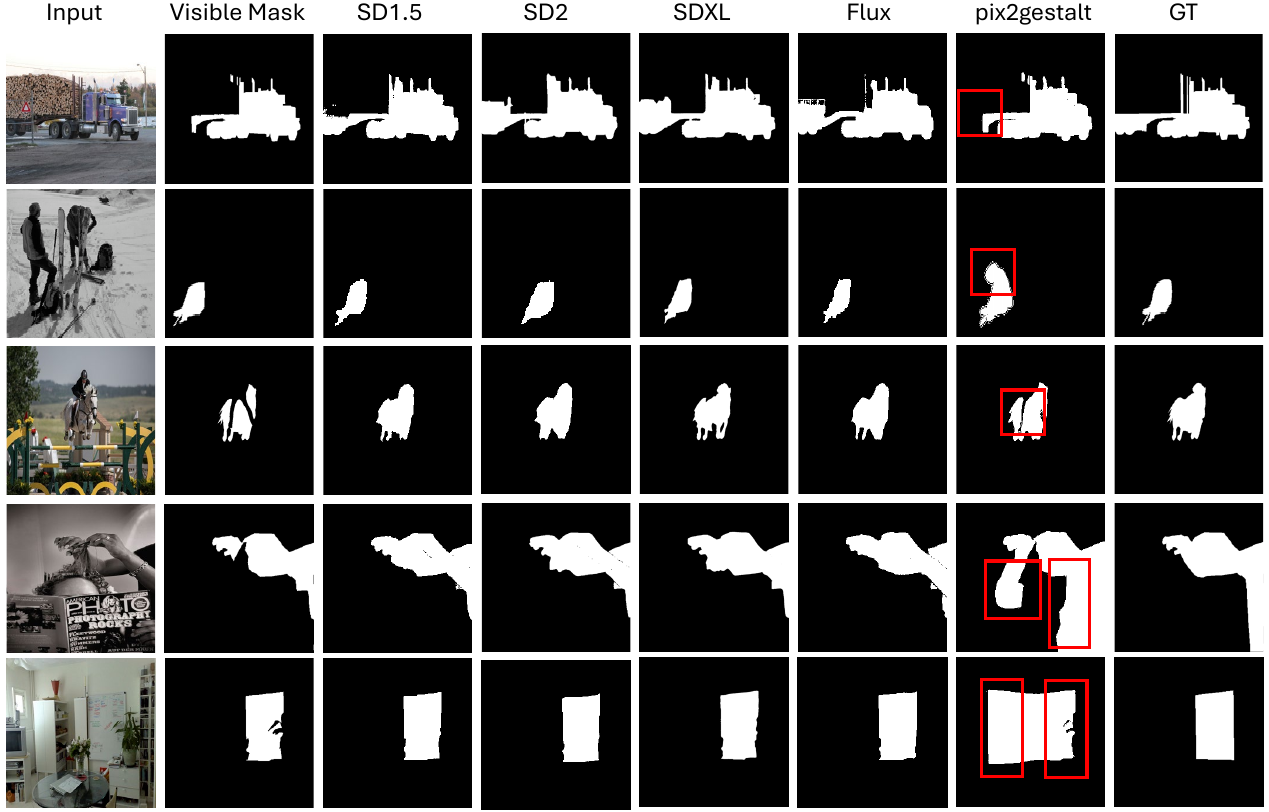}
    \caption{We show visual results of an amodal segmentation task based on COCO-A~\cite{cocoa}.}
    \label{fig:cocoa_qualitative_comp}
\end{figure}

Using COCO-A~\cite{cocoa} dataset, we show a side-by-side comparison in~\cref{fig:cocoa_qualitative_comp}. The first row shows that pix2gestalt does not generate any part from the visible mask. The second and fifth rows show that pix2gestalt overestimates compared to our works. The third row shows that pix2gestalt fails to generate an occluded region, but ours can generate the missing pixel information. 
In the fourth row, the baseline ignores the scene context, which is a stretched arm, by putting a bent arm instead, while all of our methods successfully generate the stretched arm.

\begin{figure}[ht]
    \centering
    \includegraphics[width=1.0\linewidth]
    {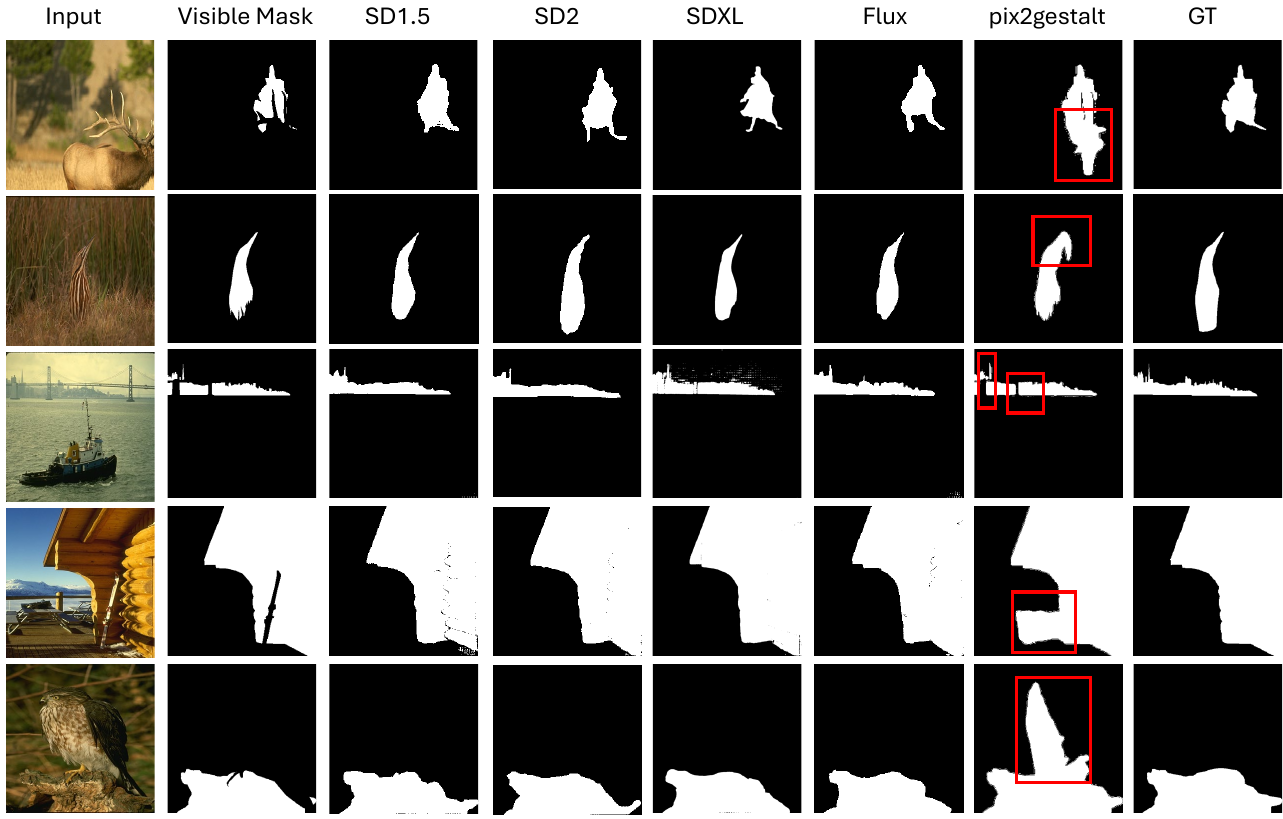}
    \caption{We show visual results of an amodal segmentation task based on BSDS-A~\cite{bsdsa}.}
    \label{fig:bsds_qualitative_comp}
\end{figure}

Similar issues from pix2gestalt are observed from BSDS-A~\cite{bsdsa} dataset, as shown in~\cref{fig:bsds_qualitative_comp}.
The first, second, fourth, and fifth rows show the recurring overestimating issue while ours can generate similar to the ground truth. The third row shows that the missing pixel information still exists, while our work can predict the occluded regions.

\begin{figure}[ht]
    \centering
    \includegraphics[width=1.0\linewidth]
    {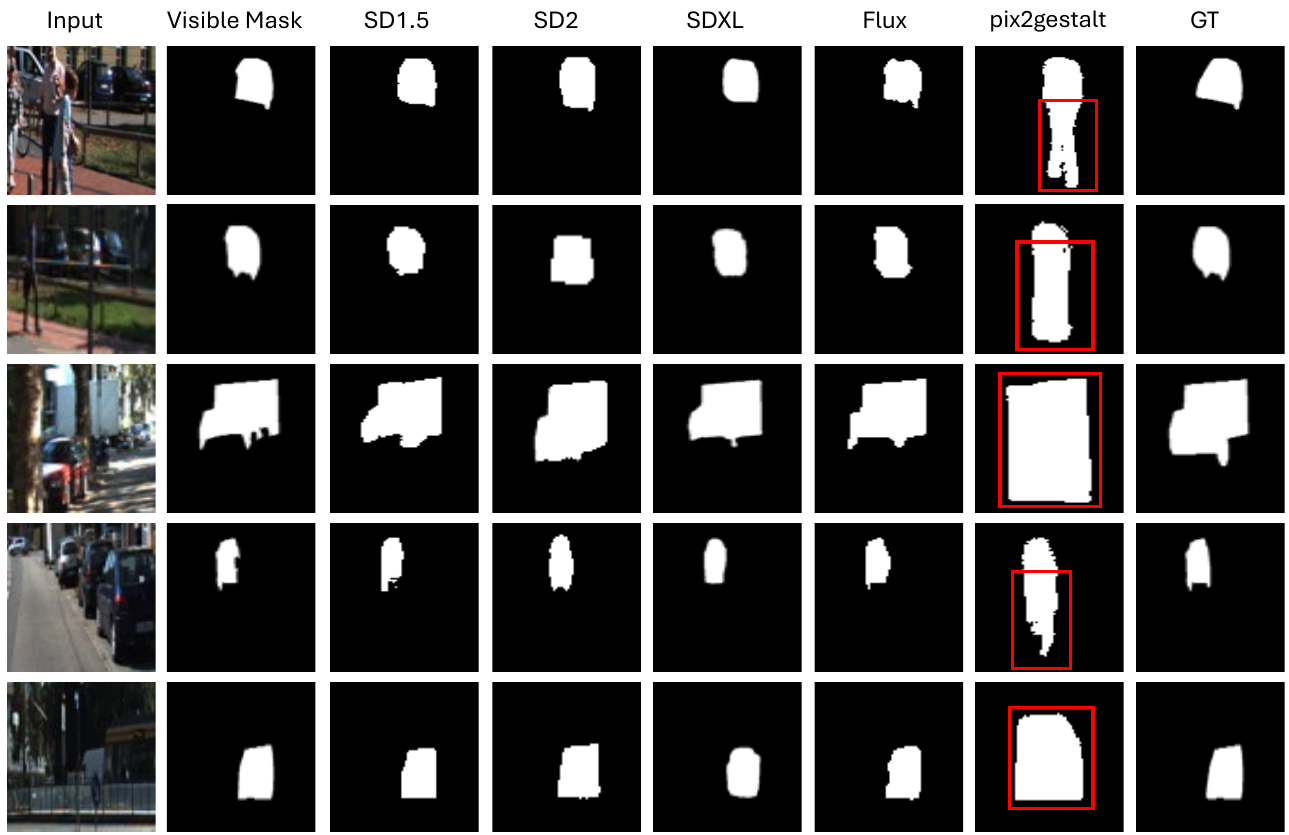}
    \caption{We show visual results of an amodal segmentation task based on KINS~\cite{kins}.}
    \label{fig:kins_qualitative_comp}
\end{figure}

In~\cref{fig:kins_qualitative_comp}, we show examples from the KINS~\cite{kins} dataset. From the third row, pix2gestalt covers most regions irrelevant to the visible mask. This dataset is not used in the original report in pix2gestalt, so the KINS dataset distribution might not be considered when constructing the training dataset, which shows the difficulties in designing customized datasets for robustness. On the other hand, our method is tuning-free and leverages the Internet-scaled pre-trained foundation models.

\begin{figure}[ht]
    \centering
    \includegraphics[width=1.0\linewidth]
    {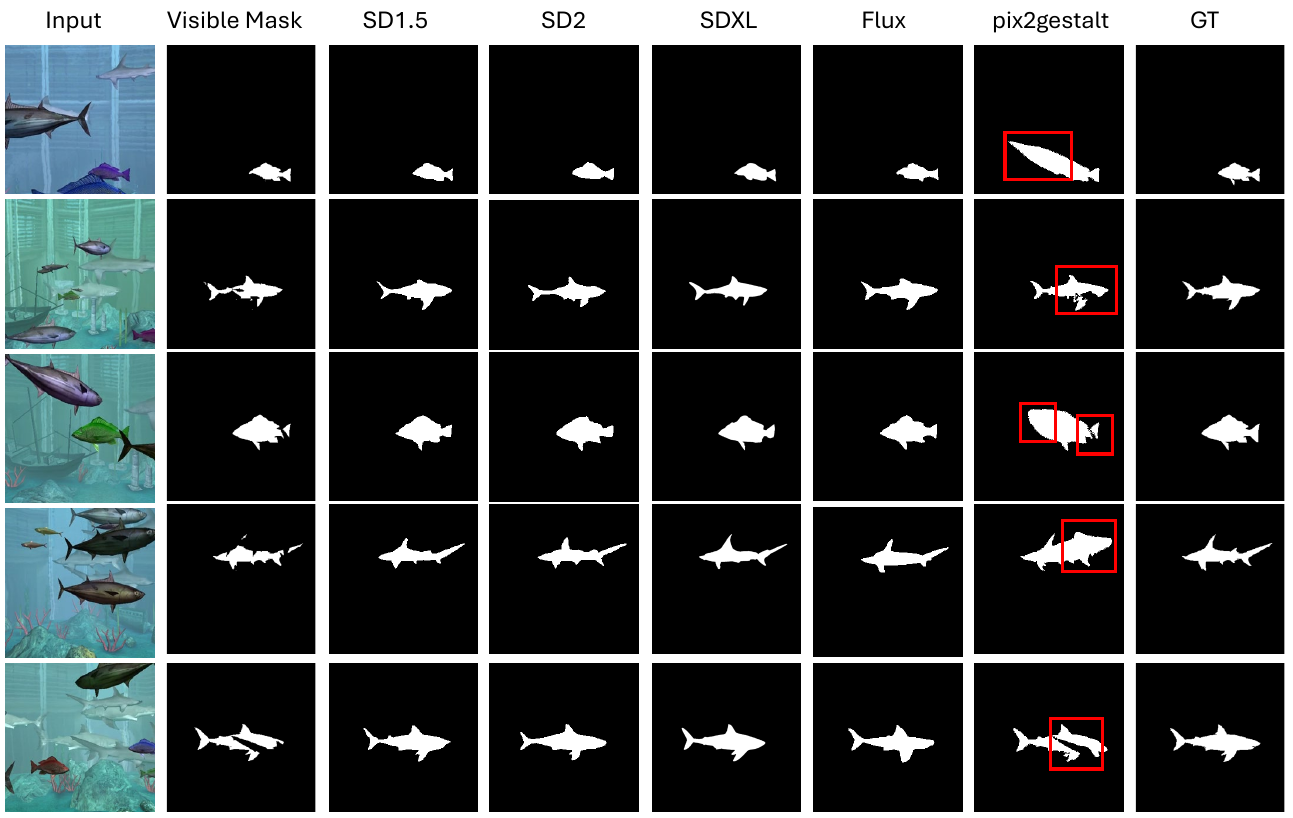}
    \caption{We show visual results of an amodal segmentation task based on FishBowl~\cite{fishbowl}.}
    \label{fig:fish_qualitative_comp}
\end{figure}

\begin{figure}[ht]
    \centering
    \includegraphics[width=1.0\linewidth]
    {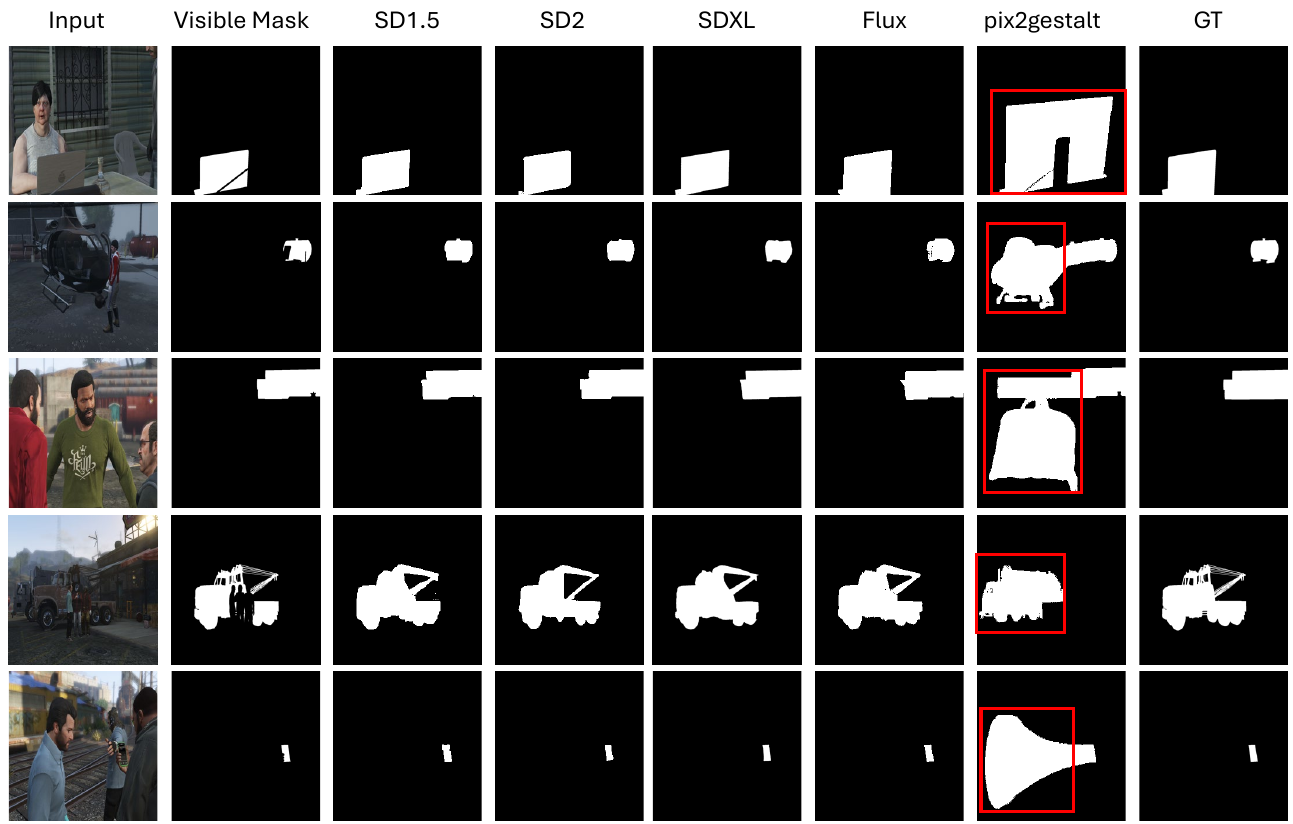}
    \caption{We show visual results of an amodal segmentation task based on SAILVOS~\cite{SAILVOS}.}
    \label{fig:sailvos_qualitative_comp}
\end{figure}

This issue remains for pix2gestalt when generating amodal masks in FishBowl~\cite{fishbowl} as shown in~\cref{fig:fish_qualitative_comp}. The first and third rows still demonstrate overgrown issues from pix2gestalt generated amodal masks. The second and fifth rows still have missing pixel information, but our methods can predict the occluded regions.
From the fourth row, when a visible mask is occluded heavily, the generated amodal mask from the baseline is a symmetrical shape. This dataset is not originally included in the report of pix2gestalt.

SAILVOS~\cite{SAILVOS} dataset reveals a new issue: hallucination.
As we show in~\cref{fig:sailvos_qualitative_comp}, all the rows show that pix2gestalt generates a random object that cannot be retrieved from the given scene. The fourth row shows the same category, which is a truck, but the details are not similar. %
While pix2gestalt constructs a customized dataset with 800k image pairs, it seems that pix2gestalt cannot cover all the occluded scenarios, while our tuning-free approach handles them well.

\section{Additional Ablation Study}
\label{sec:supp_ab}

{\bf\noindent Computation efficiency.} We study the efficiency of our approach during the amodal mask generation compared to pix2gestalt. \tabref{tab:run_stats} reports the VRAM usage and inference time. The smallest model SD2 is 4.1$\times$ more efficient in memory, and the inference (0.3 seconds) is 19$\times$ faster compared to pix2gestalt~\cite{ozguroglu2024pix2gestalt}. Additionally, our best model (SDXL) is also more efficient than pix2gestalt. 
\begin{table}[hbt]
    \setlength{\tabcolsep}{2pt}
    \centering
    \resizebox{0.99\linewidth}{!}{%
    \begin{tabular}{cccccc}
    \specialrule{.15em}{.05em}{.05em}
     & SD1.5~\cite{Rombach_2022_CVPR} & SDXL~\cite{podell2023sdxl} & SD2~\cite{SD2} & Flux~\cite{flux} & pix2gestalt~\cite{ozguroglu2024pix2gestalt}\\ %
    \midrule
    VRAM (GB) \( \downarrow \) & 3.9 & 10.6 & 3.7 & 21.5 & 15.3 \\
    Inference (s) \( \downarrow \) & 0.3 & 1.2 & 0.3 & 0.5 & 5.7 \\

    \specialrule{.15em}{.05em}{.05em}
    \end{tabular}
    }
        \vspace{-0.25cm}
    \caption{We show peak GPU VRAM usage and inference time.}
    \label{tab:run_stats}
        \vspace{-0.5cm}
\end{table}

 We first investigate further mIoU between the inpainting areas, \( \mM \), and the ground truth amodal masks to validate our method by re-purposing the inpainting models to complete the missing pixel information. In~\cref{tab:convex_miou}, we show mIoU values from each dataset in different occlusion ranges. The mIoU difference between our best values and the mask coverage from each occlusion rate of all the datasets is 21.2\%. Therefore, our proposed method generates a more accurate amodal mask than naively using the making area which takes unions of contours of visible masks as we explained.

To show that our proposed inpainting area method is effective, we create two additional mask baselines: 1) a naive bounding box style mask around the visible mask and 2) a depth-map-based mask using a deep learning-based monocular depth estimation~\cite{yang2024depth}, and those masks are presented as $R, D$, respectively. $R$ is constructed with the smallest bounding box containing the visible mask of a target object to be inpatined. To generate $D$, we only include objects that are neighbors within 10 pixel-wise L2 distances from the target object. Then, we filter out any objects farther than the target object using the estimated depth map. Intuitively, we construct $D$ that are close enough and objects that are in front of the target object, which causes an occlusion, in a certain range, $r$.

Using COCO-A~\cite{cocoa}, our proposed mask, $\mM$ method generates 3.8\% more accurate amodal masks on average from each occlusion rate as we shown in~\cref{tab:ab_cocoa}. Based on this dataset, when the occlusion rates are less than 50\% and 40\%, SD1.5~\cite{Rombach_2022_CVPR} with mask $R$ and 40\%, 30\%, 20\%, 10\%, and 5\% occlusion rates show that SDXL with mask $D$ is the second-most accurate amodal mask.
These trends continue with BSDS-A~\cite{bsdsa}. Our proposed mask, $\mM$, generates 1.6\% better quality of amodal mask instead of using two masks $R, D$ as shown in~\cref{tab:ab_bsds}. Among using the two masks with different foundation models, Flux with mask $D$ generates the second most accurate amodal mask when the objects are occluded less than 50\% and 40\%, and SDXL~\cite{podell2023sdxl} with $R$ works the second best when the occlusion rates are 30\%, 20\%, 10\% and 5\% or less.
From~\cref{tab:ab_kins} using KINS~\cite{kins}, we observed an interesting result that using a depth-aware mask, $D$, generates 5.0\% more accurate amodal mask when the occlusion rates are 50\%, 40\%, 30\%, and 20\% or less compared to the method with mask $\mM$. When the objects are occluded 10\%, 5\% or less, mask $\mM$ generates 6.1\% more accurate amodal masks.
In contrast to KINS~\cite{kins}, \cref{tab:ab_fish} shows mask $\mM$ on FishBowl~\cite{fishbowl} predicts 2.3\% more accurate amodal mask on average of six different occlusion rate ranges. The mask $\mM$ performs better when the objects are occluded 40\%, 30\%, 20\%, 10\%, and 5\% or less. Only SDXL~\cite{podell2023sdxl} with mask $D$ performs 0.2\% better when the occlusion rate is 50\% or less compared to the mask $\mM$.
\cref{tab:ab_SAILVOS} shows that mask $\mM$ generates 4.9\% better quality of accurate amodal mask on all occlusion rate ranges compared to the masks $R, D$ with other four foundation models including SD1.5~\cite{Rombach_2022_CVPR}, SD2~\cite{SD2}, SDXL~\cite{podell2023sdxl} and Flux~\cite{flux}.

\begin{table}[t]
    \setlength{\tabcolsep}{2pt}
    \centering
    \resizebox{\linewidth}{!}{%
    \begin{tabular}{ccccccc}
    \specialrule{.15em}{.05em}{.05em}
    Dataset & $\leq$50\% & $\leq$40\% & $\leq$30\% &$\leq$20\% & $\leq$10\% & $\leq$5\%\\ %
    \midrule
    COCO-A~\cite{cocoa}         & 66.5  & 67.7 & 68.8 & 69.6 & 71.4 & 71.6\\
    BSDS-A~\cite{bsdsa}         & 75.2  & 75.1 & 75.0 & 74.4 & 73.5 & 73.7\\
    KINS~\cite{kins}            & 66.7  & 68.9 & 71.4 & 74.2 & 77.8 & 80.1\\
    FishBowl~\cite{fishbowl}    & 59.4  & 59.7 & 60.0 & 60.1 & 60.4 & 60.7\\
    SAILVOS~\cite{SAILVOS}     & 30.0  & 30.2 & 30.4 & 30.8 & 31.0 & 30.9\\
    \specialrule{.15em}{.05em}{.05em}
    \end{tabular}
    }
    \caption{We evaluate mIoU(\%)$\uparrow$ with our inpainting area, \( \mM\), and ground truth amodal segmentation masks from five different datasets among various occlusion rates from 50\% to 5\%.
    } 
    \label{tab:convex_miou}
\end{table}

\begin{table}[t]
    \setlength{\tabcolsep}{2pt}
    \centering
    \resizebox{\linewidth}{!}{%
    \begin{tabular}{cccccccc}
    \specialrule{.15em}{.05em}{.05em}
    DiffMod & Mask & $\leq$50\% & $\leq$40\% & $\leq$30\% & $\leq$20\% & $\leq$10\% & $\leq$5\%\\ %
    \hline
    \rowcolor{mygray}
    Ours & $\mM$ & \textbf{82.7} & \textbf{85.4} & \textbf{88.0} & \textbf{90.6} & \textbf{93.6} & \textbf{95.0}\\
    \hline
    SD1.5 & $R$   & {\color{blue}77.7} & {\color{blue}80.0} & 82.1 & 84.1 & 86.9 & 88.2     \\
    SD1.5 & $D$   & 75.6 & 78.1 & 80.4 & 82.7 & 85.8 & 87.3     \\
    SD2 & $R$   & 75.4& 77.7 & 79.8	& 81.8 & 84.5 & 85.6    \\
    SD2 & $D$   & 75.8 & 78.2 & 80.4 & 82.7 & 85.6 & 86.9     \\
    SDXL & $R$   & 76.7 & 79.5 & 82.0 & 84.6 & 87.7 & 88.9     \\
    SDXL & $D$   & 77.0 & {\color{blue}80.0} & {\color{blue}82.5} & {\color{blue}85.1} & {\color{blue}88.6} & {\color{blue}90.6}     \\
    Flux & $R$   & 75.0 & 77.4 & 79.5 & 81.4 & 83.8 & 84.8     \\
    Flux & $D$   & 76.8 & 79.3 & 81.5 & 83.7 & 86.4 & 87.5     \\
    \specialrule{.15em}{.05em}{.05em}
    \end{tabular}
    }
        \vspace{-0.2cm}
    \caption{We evaluate our approach using COCO-A~\cite{cocoa} with the other two masks approaches, (\( R, D\)), using mIoU(\%)\(\uparrow\).}
    \label{tab:ab_cocoa}
        \vspace{-0.1cm}
\end{table}

\begin{table}[t]
    \setlength{\tabcolsep}{2pt}
    \centering
    \resizebox{\linewidth}{!}{%
    \begin{tabular}{cccccccc}
    \specialrule{.15em}{.05em}{.05em}
    DiffMod & Mask & $\leq$50\% & $\leq$40\% & $\leq$30\% & $\leq$20\% & $\leq$10\% & $\leq$5\%\\ %
    \hline
    \rowcolor{mygray}
    Ours & $\mM$ & \textbf{78.3} & \textbf{80.5} & \textbf{82.8} & \textbf{85.8} & \textbf{88.9} & \textbf{90.7}\\
    \hline
    SD1.5 & $R$   & 74.8 & 77.8 & 80.4 & 83.1 & 86.0 & 87.6      \\
    SD1.5 & $D$   & 74.7 & 77.6 & 80.3 & 83.2 & 86.2 & 87.9     \\
    SD2 & $R$   & 75.6 & 78.4 & 80.8 & 83.2	& 85.6 & 87.2   \\
    SD2 & $D$   & 75.0 & 77.5 & 79.9 & 82.4 & 84.6 & 86.2    \\
    SDXL & $R$   & 75.0 & 78.4 & {\color{blue}81.3} & {\color{blue}84.5} & {\color{blue}87.7} & {\color{blue}89.1}     \\
    SDXL & $D$   & 74.7 & 77.4 & 79.9 & 82.6 & 85.7 & 87.4     \\
    Flux & $R$   & 74.6 & 77.5 & 79.9 & 82.3 & 84.6 & 86.0     \\
    Flux & $D$   & {\color{blue}76.2} & {\color{blue}78.8} & 81.0 & 83.2 & 85.6 & 86.3     \\
    \specialrule{.15em}{.05em}{.05em}
    \end{tabular}
    }
        \vspace{-0.2cm}
    \caption{We evaluate our approach using BSDS-A~\cite{bsdsa} with the other two masks approaches, (\( R, D\)), using mIoU(\%)\(\uparrow\).}
    \label{tab:ab_bsds}
        \vspace{-0.1cm}
\end{table}

\begin{table}[t]
    \setlength{\tabcolsep}{2pt}
    \centering
    \resizebox{\linewidth}{!}{%
    \begin{tabular}{cccccccc}
    \specialrule{.15em}{.05em}{.05em}
    DiffMod & Mask & $\leq$50\% & $\leq$40\% & $\leq$30\% & $\leq$20\% & $\leq$10\% & $\leq$5\%\\ %
    \hline
    \rowcolor{mygray}
    Ours & $\mM$ & 64.8 & 66.2 & {\color{blue}68.2} & 70.6 & \textbf{74.1} & \textbf{77.3} \\
    \hline
    SD1.5 & $R$   & 60.7 & 61.8 & 63.2 & 64.9 & 67.5 & 70.7     \\
    SD1.5 & $D$   & 63.5 & 64.8 & 66.5 & 68.6 & 71.7 & 74.8    \\
    SD2 & $R$   & 61.7 & 62.7 & 63.9 & 65.4	& 67.6 & 70.7 \\
    SD2 & $D$   & 63.4& 64.4 & 65.8	& 67.5 & 69.9 & 72.9 \\
    SDXL & $R$   & 46.8 & 48.4 & 50.9 & 54.1 & 60.2 & 69.1     \\
    SDXL & $D$   & 60.4 & 62.1 & 66.6 & {\color{blue}70.8} & 67.3 & 70.5   \\
    Flux & $R$   & {\color{blue}67.5} & {\color{blue}71.7} & 60.2 & {\color{blue}70.8} & 69.8 & 73.9     \\
    Flux & $D$   & \textbf{70.7} & \textbf{74.8} & \textbf{69.1} & \textbf{75.1} & {\color{blue}72.3} & {\color{blue}77.0}    \\
    \specialrule{.15em}{.05em}{.05em}
    \end{tabular}
    }
        \vspace{-0.2cm}
    \caption{We evaluate our approach using KINS~\cite{kins} with the other two masks approaches, (\( R, D\)), using mIoU(\%)\(\uparrow\).}
    \label{tab:ab_kins}
        \vspace{-0.1cm}
\end{table}

\begin{table}[t]
    \setlength{\tabcolsep}{2pt}
    \centering
    \resizebox{\linewidth}{!}{%
    \begin{tabular}{cccccccc}
    \specialrule{.15em}{.05em}{.05em}
    DiffMod & Mask & $\leq$50\% & $\leq$40\% & $\leq$30\% & $\leq$20\% & $\leq$10\% & $\leq$5\%\\ %
    \hline
    \rowcolor{mygray}
    Ours & $\mM$ & {\color{blue}80.4} & \textbf{82.8} & \textbf{85.3} & \textbf{88.0} & \textbf{90.7} & \textbf{91.9} \\
    \hline
    SD1.5 & $R$   & 60.9 & 63.0 & 65.0 & 67.3 & 70.4 & 74.3     \\
    SD1.5 & $D$   & 79.7 & 81.1 & 82.9 & 84.4 & 85.9 & 86.3    \\
    SD2 & $R$   & 72.5 & 74.6 & 76.6 & 78.6	& 80.4 & 81.1 \\
    SD2 & $D$   & 74.1 & 76.3 & 78.6 & 81.1 & 83.4 & 84.7 \\
    SDXL & $R$   & 76.9 & 79.2 & 81.6 & 83.8 & 85.8 & 86.7     \\
    SDXL & $D$   & \textbf{80.6} & {\color{blue}82.1} & {\color{blue}84.0} & {\color{blue}85.3} & {\color{blue}86.8} & {\color{blue}86.8}   \\
    Flux & $R$   & 77.6 & 78.9 & 80.3 & 81.3 & 82.0 & 81.9     \\
    Flux & $D$   & 64.1 & 65.4 & 67.3 & 69.5 & 72.7 & 75.6    \\
    \specialrule{.15em}{.05em}{.05em}
    \end{tabular}
    }
        \vspace{-0.2cm}
    \caption{We evaluate our approach using FishBowl~\cite{fei2024dysen} with the other two masks approaches, (\( R, D\)), using mIoU(\%)\(\uparrow\).}
    \label{tab:ab_fish}
        \vspace{-0.1cm}
\end{table}

\begin{table}[t]
    \setlength{\tabcolsep}{2pt}
    \centering
    \resizebox{\linewidth}{!}{%
    \begin{tabular}{cccccccc}
    \specialrule{.15em}{.05em}{.05em}
    DiffMod & Mask & $\leq$50\% & $\leq$40\% & $\leq$30\% & $\leq$20\% & $\leq$10\% & $\leq$5\%\\ %
    \hline
    \rowcolor{mygray}
    Ours & $\mM$ & \textbf{80.0} & \textbf{82.3} & \textbf{84.1} & \textbf{85.8} & \textbf{87.4} & \textbf{88.8} \\
    \hline
    SD1.5 & $R$   & 67.5 & 69.4 & 70.8 & 72.2 & 74.0 & 76.1     \\
    SD1.5 & $D$   & 65.6 & 67.4 & 68.8 & 70.5 & 72.3 & 74.6    \\
    SD2 & $R$   & 68.2 & 69.8 & 71.2 & 72.7 & 75.5 & 77.3 \\
    SD2 & $D$   & 69.7& 71.4 & 72.8	& 74.5 & 77.1 & 78.8 \\
    SDXL & $R$   & 57.4 & 59.0 & 60.6 & 62.5 & 65.6 & 67.6     \\
    SDXL & $D$   & 63.5 & 65.1 & 66.6 & 68.2 & 70.7 & 72.3    \\
    Flux & $R$   & 66.7 & 68.6 & 70.1 & 71.7 & 73.4 & 75.8     \\
    Flux & $D$   & {\color{blue}76.1} & {\color{blue}77.8} & {\color{blue}79.2} & {\color{blue}80.5} & {\color{blue}81.9} & {\color{blue}83.6}    \\
    \specialrule{.15em}{.05em}{.05em}
    \end{tabular}
    }
        \vspace{-0.2cm}
    \caption{We evaluate our approach using SAILVOS~\cite{SAILVOS} with the other two masks approaches, (\( R, D\)), using mIoU(\%)\(\uparrow\).}
    \label{tab:ab_SAILVOS}
        \vspace{-0.1cm}
\end{table}

\section{Implementation Details}
\label{sec:supp_implementation}
Our implementation is based on Diffusers library version \textsf{0.31.0.dev0} from \url{https://github.com/huggingface/diffusers} and Flux inpainting from \url{https://github.com/Gothos/diffusers/tree/flux-inpaint} using Pytorch \textsf{2.4.1+cu124} and diffusion inpainting models from Hugging Face. The URL for SD15, SDXL, SD2 and Flux are
\begin{itemize}
    \item \url{https://huggingface.co/benjamin-paine/stable-diffusion-v1-5-inpainting}, 
    \item \url{https://huggingface.co/diffusers/stable-diffusion-xl-1.0-inpainting-0.1}, 
    \item \url{https://huggingface.co/stabilityai/stable-diffusion-2-inpainting}
    \item \url{https://huggingface.co/black-forest-labs/FLUX.1-schnell} 
\end{itemize}
To utilize Hugging face diffusion-based inpainting models, we use \textsf{AutoPipelineForInpainting} module from the Diffusers library. 
Especially, we set \textsf{num\_inference\_steps} as 20, which is not a high value, since our goal is not generating photo-realistic quality images.
All foundation models can fit into 24GB VRAM GPU, but we are unable to load Flux~\cite{flux}, so we apply 8-bit quantization using \textsf{optimum-quanto} from \url{https://github.com/huggingface/optimum-quanto}.
We will release our implementation upon acceptance of this work.

\end{document}